\documentclass{article}

\usepackage{iclr2027_conference,times}
\usepackage[utf8]{inputenc}
\usepackage{amsmath, amssymb}
\usepackage{microtype}

\usepackage{xcolor}
\usepackage{tcolorbox}
\tcbuselibrary{skins, breakable}

\usepackage{pifont}

\usepackage{graphicx}
\usepackage{hyperref}
\hypersetup{hidelinks}
\usepackage{subcaption}
\usepackage{booktabs}
\usepackage{tabularx}
\usepackage{array}

\newcommand{\appendixtablestyle}{%
  \scriptsize
  \setlength{\tabcolsep}{4pt}%
  \renewcommand{\arraystretch}{1.04}%
}
\newcolumntype{Y}{>{\raggedright\arraybackslash}X}

\usepackage{multirow}
\usepackage{float}

\usepackage{amsmath,amsfonts,bm}

\def\secref#1{section~\ref{#1}}
\def\Secref#1{Section~\ref{#1}}

\def\eqref#1{equation~\ref{#1}}
\def\Eqref#1{Equation~\ref{#1}}

\def\algref#1{algorithm~\ref{#1}}
\def\Algref#1{Algorithm~\ref{#1}}

\def\ceil#1{\lceil #1 \rceil}

\def\1{\bm{1}}

\def\mG{{\bm{G}}}
\def\mH{{\bm{H}}}

\def\mW{{\bm{W}}}

\def\mY{{\bm{Y}}}

\DeclareMathAlphabet{\mathsfit}{\encodingdefault}{\sfdefault}{m}{sl}
\SetMathAlphabet{\mathsfit}{bold}{\encodingdefault}{\sfdefault}{bx}{n}

\newcommand{\E}{\mathbb{E}}
\newcommand{\Ls}{\mathcal{L}}

\DeclareMathOperator{\Mean}{Mean}
\newcommand{\round}[1]{\left\lfloor #1 \right\rceil}

\floatstyle{ruled}
\newfloat{algorithm}{tbp}{loa}
\floatname{algorithm}{Algorithm}
\newenvironment{algorithmic}[1][]{%
  \begin{minipage}{\linewidth}\small
}{%
  \end{minipage}
}
\newcommand{\Require}{\par\noindent\textbf{Require:}~}
\newcommand{\Ensure}{\par\noindent\textbf{Ensure:}~}
\newcommand{\State}{\par\noindent}
\newcommand{\For}[1]{%
  \par\noindent\textbf{for} #1 \textbf{do}\par
  \begingroup\leftskip=1em
}
\newcommand{\EndFor}{%
  \par\endgroup\noindent\textbf{end for}
}
\newcommand{\Return}{\textbf{return}~}

\newcommand{\resna}{\textemdash}

\title{Relative Kinetic Utility: Calibrating Cross-Layer Credit for Global Structured LLM Pruning}

\iclrfinalcopy

\author{
Tianhao Qian$^{1}$ \quad
Guilin Qi$^{1,*}$ \quad
Jiayu Chen$^{2,*}$ \\[3pt]
\normalfont\small
$^{1}$Southeast University
\qquad
$^{2}$The University of Hong Kong \\[2pt]
\normalfont\small
$^{*}$Corresponding authors
}

\date{}

\begin{document}
\raggedbottom

\maketitle
\thispagestyle{empty}

\begin{abstract}
Global structured pruning requires channels from different layers to compete
under a shared sparsity budget, raising two coupled challenges: identifying which channels should be retained and making their scores comparable across layers. Raw channel scores can contain block-common scale
that leaves within-block ordering unchanged but distorts model-wide competition. Our experiment indicates that similar layer-wise allocations can retain substantially different FFN channels, so layer allocation alone does not determine channel identity. Motivated by this separation, we introduce \textsc{Global Relative Kinetic Utility} (Global RKU), a label-free criterion that separates channel importance estimation from cross-layer comparison. Global RKU measures channel participation using a final-hidden-state activation--gradient signal, then applies block-relative normalization to mitigate block-common scale while preserving within-block ordering, requires only unlabeled calibration inputs, and produces a static pruning topology in a single calibration stage. Under questions-only calibration on Qwen-2.5-7B, RKU--GISP Mean3 margins are $-0.98/+3.79/+8.61$ at 30/40/50\% sparsity (average $+3.81$). Additional Qwen evaluations cover non-mathematical reasoning, recovery, held-out transfer, and physical deployment. Separately, replacing Wiki16K with questions-only Q16K improves RKU's Mean3 at every tested sparsity on Qwen, Llama, and Gemma. Our ablation study shows relative-normalization gains of 14.42/5.53 Mean3 points at 40/50\% sparsity; the common-seed audit is positive in all 27 seed--task comparisons.

\end{abstract}

\section{Introduction}
\label{sec}

Post-training pruning removes parameters or structures from a pretrained LLM
without repeating pre-training, and can be broadly divided into unstructured
and structured pruning
\cite{frantar2023sparsegpt,sun2024wanda,ma2023llm}. Unstructured pruning can
preserve model quality at high sparsity, but its irregular sparsity patterns
often require specialized kernels or hardware to realize practical speedups.
Structured pruning instead removes complete heads, channels, dimensions, or
blocks, producing smaller dense operators that can directly reduce memory and
computation
\cite{an2024flap,ashkboos2024slicegpt,xia2024sheared,gao2024displlm,men2025shortgpt}. A central design choice in structured pruning is how the pruning budget is distributed across layers.
Early structured methods often apply similar sparsity ratios across layers,
whereas recent work increasingly allocates capacity non-uniformly across the model
\cite{yin2024owl,xu2024besa,li2024dsa,bai2024sparsellm,li2025tyr,he2025olica,yang2025tell}. Under a shared model-wide budget, structures from different blocks must then compete for retention. 

This model-wide competition creates two distinct problems. The first is channel-level \emph{credit assignment}, which determines what evidence should justify retaining an individual channel. Existing methods derive structural
importance from activation statistics, reconstruction error, loss sensitivity, functional grouping, learned gates, or adaptive search
\cite{an2024flap,shin2024rethink,wei2024sobp,li2024dsa,yu2026fang,wang2026gisp,wang2026grasprune}. The second is \emph{cross-layer comparison}, which determines how channel
scores should be compared when channels from different blocks compete for the same budget. Recent global and non-uniform pruning methods address model-wide allocation through normalized importance, learned gates, or explicit
optimization
\cite{li2024dsa,bai2024sparsellm,wang2026gisp,wang2026grasprune}. In several representative global or non-uniform methods, credit estimation and cross-layer allocation are coupled inside the same scoring or optimization procedure. This coupling matters because a multiplicative factor shared by all channels in one block can leave
their within-block ordering unchanged while changing how much capacity that block receives globally. Thus a score that is useful for ranking channels within a block is not necessarily comparable across blocks. Calibration data
and optimization objectives can further change the resulting scores and masks
\cite{bandari2024c4,ji2025calibration,kurz2026calibrationlanguage,wang2026gisp}.

Our controlled interventions reveal that layer allocation does not determine channel
identity. For FANG \cite{yu2026fang}, all 32 layers retain identical channel counts across
calibration corpora, yet FFN-mask Jaccard is only $0.674$--$0.696$. For GISP \cite{wang2026gisp},
layer-wise allocation remains highly correlated (Spearman $0.967$--$0.981$), while mask Jaccard falls to $0.651$--$0.690$
(Appendix Table~\ref{tab:app_credit_motivation}). Thus, nearly unchanged layer allocation can hide substantial changes in which
channels survive, motivating a distinction between within-block channel credit and the scale governing cross-block competition.

To separate these two roles, we propose \textsc{Global Relative Kinetic Utility} (Global RKU). Inspired by AGF's amplitude-based view of feature utility \cite{kunin2025alternating}, RKU assigns functional retention credit to each active SwiGLU channel using a final-hidden-state activation--gradient participation signal: AGF evaluates a dormant feature by increasing its amplitude from zero, whereas RKU measures the effect of rescaling a feature that is already active. Before channels compete under the model-wide budget, RKU divides each channel's participation by the aggregate participation of its block. We call the multiplicative factor shared within a block \emph{block-common scale}; under approximate block-scale factorization, relative normalization removes this scale while preserving within-block ordering. Here, \emph{label-free} means no target answers, solution traces, or target-loss labels; RKU then produces one static topology without gate optimization or iterative pruning. Keeping the participation signal, calibration data, pruning unit, global budget, and selection fixed, relative normalization improves Mean3 by 14.42/5.53 points at 40/50\% sparsity in the signal-factorial ablation (Table~\ref{tab:main_signal_factorial}); all common-seed margins are positive (Table~\ref{tab:main_global_step0_seed_deltas}). Under questions-only calibration on Qwen-2.5-7B, RKU--GISP Mean3 margins are $-0.98/+3.79/+8.61$ at 30/40/50\% sparsity (average $+3.81$). Separately, replacing Wiki16K with questions-only Q16K improves RKU's Mean3 at every tested sparsity on Qwen, Llama, and Gemma. Among the three matched FFN-only methods evaluated on the non-mathematical tasks, RKU has the highest mean accuracy in eight of nine cells, including all three tasks at 50\% sparsity. The Qwen experiments also report held-out transfer and physical-deployment measurements.

\textbf{Our contributions are as follows:}
\begin{itemize}

\item \textbf{Functional retention credit.}
We formulate channel-level credit assignment as a separate problem in global structured FFN pruning and introduce a label-free activation--gradient participation criterion for measuring how active channels contribute to later frozen-model computation.

\item \textbf{Explicit separation of channel credit and cross-layer calibration.}
We formulate channel-level credit estimation and cross-layer score calibration as distinct stages of global structured pruning. For RKU's activation--gradient participation signal, block-relative normalization removes a block-common multiplicative component under approximate block-scale factorization while preserving within-block ordering. The signal-factorial ablation isolates this stage, yielding +14.42/+5.53 Mean3 points at 40/50\% sparsity; all 27 common-seed task margins are positive (Tables~\ref{tab:main_signal_factorial} and~\ref{tab:main_global_step0_seed_deltas}).

\item \textbf{Empirical validation.}
Under Q16K calibration on Qwen-2.5-7B, RKU--GISP Mean3 margins are
$-0.98/+3.79/+8.61$ at 30/40/50\% sparsity (average $+3.81$).
For RKU, Q16K improves over Wiki16K at every tested sparsity on Qwen, Llama,
and Gemma. Among the matched FFN-only methods, RKU has the highest mean accuracy
in eight of nine non-mathematical cells. We also report held-out MathQA transfer,
recovery, and physical-deployment measurements on Qwen.

\end{itemize}

\section{Related Work}
\label{sec:related_work}

\paragraph{Structured Pruning and Global Allocation.}
Structured pruning removes attention heads, FFN channels, hidden dimensions,
or blocks to obtain compact dense models
\cite{michel2019heads,voita2019heads,lagunas2021block,xia2022cofi,
ashkboos2024slicegpt,xia2024sheared,kurtic2023ziplm,gao2024displlm,men2025shortgpt}.
Recent methods relax uniform layer-wise pruning by allocating capacity
non-uniformly across the network. FLAP constructs adaptive layer widths from
standardized structural importance \cite{an2024flap}; DSA searches for
layer-wise sparsity allocations \cite{li2024dsa}; SparseLLM decomposes a global
pruning objective \cite{bai2024sparsellm}; GISP performs iterative global
pruning using normalized loss-based importance \cite{wang2026gisp}; GRASPrune
learns structural gates under a global budget \cite{wang2026grasprune}; and
other work searches or predicts non-uniform pruning allocations
\cite{li2025tyr,ai2025nirvana,guo2025slimllm,he2025olica,yang2025tell}.
Recent work also studies compensation under limited calibration,
reconstruction-based layer allocation, training-aware evolutionary search,
analytical global structured pruning, and calibration-aware global-budget
allocation \cite{haruta2026rcpu,yang2026lsa,tang2026darwinlm,
song2025structprune,liu2026gprunellm}. RKU studies a narrower question: how to
compare one label-free activation--gradient channel score across blocks. It
therefore treats the score and its block-relative calibration as separate stages.

\paragraph{Importance Estimation and Calibration.}
Existing pruning criteria use weight magnitude, activations, local
reconstruction error, or first-order loss sensitivity
\cite{sun2024wanda,frantar2023sparsegpt,an2024flap,
lee2019snip,molchanov2019importance,ma2023llm,wang2026gisp,zhang2024loraprune}.
NIRVANA likewise uses first-order weight--gradient saliency
\cite{ai2025nirvana}, whereas RKU measures the contribution of an active FFN
channel to later computation and converts it into a relative cross-layer
score. Calibration data can itself have a substantial effect on post-training
pruning \cite{ji2025calibration}, so we vary calibration content explicitly.
Recent pruning methods explore generic, task-specific, or self-generated calibration signals
\cite{wang2026gisp,ai2025nirvana,wang2025resp,xiang2025ssgr,ding2026lrmpruning}. RKU requires no downstream labels, target answers, or solution traces and can operate with either generic or task-input calibration text. Its final topology is fixed offline, in contrast to input-adaptive or
instruction-conditioned pruning \cite{le2025probe,ji2026ocp,hou2025instruction}.

\section{Preliminaries}
\label{sec:preliminaries}

\paragraph{Problem Setting.}
Consider a frozen decoder with $L$ Transformer blocks. In block $l$, a
SwiGLU channel $i=(l,c)$ is the coupled structural unit
$\{\mW_{\mathrm{gate}}^{(l)}[c,:],
\mW_{\mathrm{up}}^{(l)}[c,:],
\mW_{\mathrm{down}}^{(l)}[:,c]\}$. Pruning the channel removes the
same intermediate coordinate from all three projections. Let $\mathcal I_l$
be the channels in block $l$ and $\mathcal I=\cup_l\mathcal I_l$. At target
sparsity $s$, the retained-channel count is
\begin{equation}
K=\round{(1-s)|\mathcal I|}.
\label{eq:global_budget}
\end{equation}
Using only unlabeled calibration text, our goal is to assign each channel a
retention credit and select a retained set $R\in\mathcal F_K$, where
$\mathcal F_K$ contains the $K$-channel masks satisfying a fixed minimum
retained-width constraint in every block. The same constraint is used for
Global RKU and its RKU w/o Relative Calibration; its fixed value and mask
construction are specified in Appendix~\ref{app:implementation_concordance}. The problem has two distinct parts. The first is to define what
channel-level retention credit should measure. The second is to make that credit
suitable for competition across blocks under one model-wide budget. Under fixed
per-layer quotas, a block-wide rescaling cannot change the local top-$k$ set, while
under a shared budget, the same rescaling can change how much width a block
receives.

\section{Methodology}
\label{sec:methodology}
\begingroup
\setlength{\abovedisplayskip}{4pt plus 1pt minus 1pt}
\setlength{\belowdisplayskip}{4pt plus 1pt minus 1pt}
\setlength{\abovedisplayshortskip}{2pt plus 1pt minus 1pt}
\setlength{\belowdisplayshortskip}{3pt plus 1pt minus 1pt}

Global RKU addresses the two parts of the problem in sequence. It first defines channel-level retention credit using an activation--gradient signal that measures the downstream sensitivity associated with an active FFN channel. It then mitigates block-common scale in
this participation signal before channels compete under one model-wide budget.
\Algref{alg:global_rku} gives the complete procedure.

\subsection{Directional Participation}
\label{subsec:hidden_state_kinetic}

AGF defines the utility of a dormant candidate feature by its correlation
with the residual of the currently active model \citep{kunin2025alternating}:
\begin{equation}
\mathcal U_i(\theta;r)
=
\mathbb E_x
\left[
\left\langle
f_i(x;\theta),
r(x)
\right\rangle
\right],
\qquad
r(x)
=
y(x)-f(x;\Theta_{\mathcal A}).
\label{eq:agf_utility}
\end{equation}
Here, $\Theta_{\mathcal A}$ denotes the active neurons. Near an
active--dormant saddle point, AGF shows that the dynamics of a dormant
neuron are approximately driven by the gradient of this utility. Under
squared loss, the same utility can be written as a directional derivative
with respect to the amplitude of the candidate feature. Introducing a
scalar amplitude $\lambda$,
\begin{equation}
f(x;\Theta_{\mathcal A})
\longrightarrow
f(x;\Theta_{\mathcal A})
+
\lambda f_i(x;\theta),
\end{equation}
gives
\begin{equation}
-
\left.
\frac{\partial}{\partial \lambda}
\Ls\!\left(
f_{\mathcal A}+\lambda f_i
\right)
\right|_{\lambda=0}
=
\mathcal U_i(\theta;r).
\label{eq:agf_amplitude_view}
\end{equation}
Thus, AGF evaluates a dormant feature through the first-order effect of
increasing its amplitude from zero.

RKU transfers this amplitude-based view from dormant-feature introduction to activation--gradient retention. Rather than introducing a dormant feature into a supervised
predictor, we perturb the amplitude of an already active FFN channel in a
frozen language model and measure its effect on a downstream representation without using labels or task supervision. We refer to this active-channel amplitude participation as \emph{kinetic}.

For calibration input $x\in\mathcal D$, let
$\mH_{\mathrm{post}}^{(L)}(x)$ denote the hidden state after the backbone's
terminal RMSNorm, and define
\begin{equation}
\mathcal E(x)
=
\left\|
\mH_{\mathrm{post}}^{(L)}(x)
\right\|_F^2.
\label{eq:hidden_state_energy}
\end{equation}
Unlike the supervised loss underlying \Eqref{eq:agf_utility}, this probe
requires no target answers, solution traces, or downstream task loss.

Let $\mY_{l,c}(x)$ denote the activation of channel $c$ at the input to
the down projection in block $l$, and define
\begin{equation}
\mG_{l,c}(x)
=
\nabla_{\mY_{l,c}(x)}
\mathcal E(x).
\label{eq:channel_gradient}
\end{equation}
We expose the amplitude of this active channel by introducing a scalar gate
$\gamma_{l,c}$:
\begin{equation}
\mY_{l,c}(x)
\longrightarrow
\gamma_{l,c}\mY_{l,c}(x).
\end{equation}
The corresponding first-order amplitude sensitivity is
\begin{equation}
\left.
\frac{\partial \mathcal E(x)}
{\partial \log \gamma_{l,c}}
\right|_{\gamma_{l,c}=1}
=
\left\langle
\mY_{l,c}(x),
\mG_{l,c}(x)
\right\rangle.
\label{eq:gate_sensitivity}
\end{equation}
Equation~\ref{eq:gate_sensitivity} is the active-channel analogue of the
amplitude view in \Eqref{eq:agf_amplitude_view}: AGF measures the first-order
utility of increasing a dormant feature from zero, whereas RKU measures the
first-order sensitivity to rescaling a feature that is already active.

The signed inner product in \Eqref{eq:gate_sensitivity}, however, can cancel
contributions across tokens and coordinates. For pruning, we therefore retain
the coordinatewise activation--gradient interaction before aggregation and
define the nonnegative participation statistic
\begin{equation}
\mathcal A_{l,c}
=
\frac{1}{|\mathcal D|}
\sum_{x\in\mathcal D}
\Mean\!\left(
\left|
\mY_{l,c}(x)
\odot
\mG_{l,c}(x)
\right|
\right).
\label{eq:kinetic_participation}
\end{equation}
Here, $\Mean$ averages over the entries represented by $x$ (token positions
and batch entries when batched). The absolute product prevents coordinatewise
cancellation, so $\mathcal A_{l,c}$ measures the magnitude of coordinatewise activation--gradient interaction rather than signed first-order sensitivity. We use $\mathcal A_{l,c}$ as the channel-specific
retention signal in RKU. It incorporates downstream sensitivity while remaining
label-free; the final RKU score separately incorporates the down-projection norm
as a weight-magnitude factor.

\subsection{Relative Cross-Layer Credit}
\label{subsec:global_rku}

Activation--gradient participation provides the channel-specific signal used by RKU, but
its raw magnitude is not automatically suitable for model-wide competition. A
multiplicative factor shared by all channels in block $l$ leaves their within-block
ordering unchanged while changing the amount of capacity assigned to that block
under a shared budget. To separate block-common scale from channel-specific
participation, suppose that $\mathcal A_{l,c}\approx\alpha_l q_{l,c}$, where
$\alpha_l$ is a block-common factor and $q_{l,c}$ is the channel-specific
component. In the exact case,
\[
\frac{\mathcal A_{l,c}}{\sum_j\mathcal A_{l,j}}
=
\frac{q_{l,c}}{\sum_j q_{l,j}},
\]
so the common multiplier disappears whenever the block aggregate is above
the numerical floor. We therefore set
\[
Z_l=
\max\!\left\{
\sum_j \mathcal A_{l,j},
10^{-12}
\right\},
\]
and define
$w_{l,c}=\|\mW_{\mathrm{down}}^{(l)}[:,c]\|_2$. The fixed floor
$10^{-12}$ is used only to avoid division by a numerically vanishing block
aggregate. We score channel $(l,c)$ by
\begin{equation}
\mathcal S_{l,c}^{\mathrm{RKU}}
=
w_{l,c}
\frac{\mathcal A_{l,c}}{Z_l}.
\label{eq:global_rku_score}
\end{equation}
When the numerical floor does not alter the denominator, the normalized term cancels the block-common scale in the participation signal before cross-layer comparison. The relative normalization acts on $\mathcal A_{l,c}$, while the down-projection norm provides a separate weight-magnitude factor in the final
score.

A perturbation analysis for bounded channel-wise deviations from exact block-scale factorization is given in Appendix~\ref{app:method_derivations}. \Secref{subsec:signal_factorial_main} compares the normalized score with its unnormalized counterpart.

Channels are ranked by $\mathcal S_{l,c}^{\mathrm{RKU}}$, and the retained set
$R_{\mathrm{RKU}}$ is constructed under the global budget $K$ and the
per-block minimum-width constraint using Algorithm~\ref{alg:global_rku}.
The same scoring rule is used once per calibration setting, and the resulting
static mask is reused across downstream evaluations.

\begin{algorithm}[H]
\caption{Global Relative Kinetic Utility for Structured FFN Pruning}
\label{alg:global_rku}
\begin{algorithmic}[1]
\footnotesize
\Require Frozen model $f_{\Theta}$, unlabeled calibration set $\mathcal{D}$,
target global sparsity $s$, minimum retained ratio $\rho_{\min}$
\Ensure Structured FFN mask $\mathcal{M}$

\State Initialize $\mathcal{A}_{l,c}\leftarrow 0$ for every FFN channel
$(l,c)$

\For{each calibration input $x\in\mathcal{D}$}
    \State Run a forward pass and collect down-projection input
    activations $\mY_{l,c}(x)$
    \State Compute $\mathcal{E}(x)=\|\mH_{\mathrm{post}}^{(L)}(x)\|_F^2$
    \State Backpropagate
    $\mG_{l,c}(x)=\nabla_{\mY_{l,c}(x)}\mathcal{E}(x)$
    \State Accumulate
    \[
    \mathcal{A}_{l,c}
    \leftarrow
    \mathcal{A}_{l,c}
    +
    \Mean\!\left(
    |\mY_{l,c}(x)\odot\mG_{l,c}(x)|
    \right)
    \]
\EndFor

\For{each FFN channel $(l,c)$}
    \State Set $\mathcal A_{l,c}\leftarrow\mathcal A_{l,c}/|\mathcal D|$
\EndFor

\For{each Transformer block $l$}
    \State Compute
    $Z_l=\max\!\left\{\sum_j\mathcal{A}_{l,j},10^{-12}\right\}$
    \For{each FFN channel $c$}
        \State Set
        $w_{l,c}\leftarrow\|\mW_{\mathrm{down}}^{(l)}[:,c]\|_2$
        \State Compute
        \[
        \mathcal{S}_{l,c}^{\mathrm{RKU}}
        =
        w_{l,c}
        \frac{\mathcal{A}_{l,c}}{Z_l}
        \]
    \EndFor
\EndFor

\State Set $K=\round{(1-s)\sum_l d_{\mathrm{ff}}^{(l)}}$
\State Let $R$ be the globally highest-scoring $K$ channels
\State Enforce $|R\cap\mathcal I_l|\ge
\ceil{\rho_{\min}|\mathcal I_l|}$ for every block $l$
\State Remove an equal number of the lowest-scoring eligible channels from
other blocks to restore the exact global budget $K$
\State Apply each retained index jointly to the gate, up, and down projections
\State \Return $\mathcal{M}$
\end{algorithmic}
\end{algorithm}

\Algref{alg:global_rku} aggregates $\mathcal A_{l,c}$ over the full
calibration set before normalizing by $Z_l$. Each retained index is applied
jointly to the gate, up, and down projections, so one mask entry corresponds to
one coupled SwiGLU channel. For a fixed calibration setting, RKU produces one
static mask without updating model parameters or using downstream labels,
answers, or solution traces. Scoring uses one forward and one backward pass per
calibration batch, and the channel statistics can be accumulated online.

\endgroup

\section{Experiments}
\label{sec:experiments}

\paragraph{Evaluation Protocol.}
We evaluate Qwen-2.5-7B at 30\%, 40\%, and 50\% structured FFN sparsity on GSM8K, AQuA, and MathQA \cite{cobbe2021gsm8k,ling2017aqua,amini2019mathqa}, with Llama-3.1-8B and Gemma-2-9B for cross-family validation. Unless stated otherwise, results are mean $\pm$ sample SD over structural seeds 42, 100, and 2026. Recovery uses separate LoRA \cite{hu2022lora} runs on GSM8K and AQuA; paired held-out transfer is the mean of the corresponding MathQA accuracies. Throughout, Mean3 denotes the arithmetic mean of GSM8K, AQuA, and MathQA accuracies computed within each structural seed; Mean3 values are then summarized across seeds 42, 100, and 2026 as mean $\pm$ sample SD. Mathematical sweeps use structured zero masks, while the deployment audit materializes the same retained topology as compact dense submodels. The broad comparison uses Wiki16K for RKU, C4-16K for GISP, and BookCorpus for NIRVANA. Direct RKU--GISP comparisons additionally match task-input calibration content, token budget, and downstream evaluator while retaining each method's native structural scope: RKU prunes coupled FFN channels only, whereas GISP globally competes over both MLP channels and GQA-compatible attention groups. RKU w/o Relative Calibration keeps the participation signal, pruning unit, global budget, minimum retained-width constraint, selection protocol, and calibration data fixed while omitting $Z_l$. The appendices specify structured-control definitions, seed semantics,
calibration construction, recovery hyperparameters, hardware settings,
physical-equivalence checks, and seed-level diagnostics in
Appendices~\ref{app:resource_accounting}, \ref{app:llama_complete_results},
and~\ref{app:credit_diagnostic}.

\paragraph{Allocation--Identity Decoupling.}
We test the distinction motivating RKU by perturbing the calibration
corpus or preservation objective of three representative structured-pruning
methods. Across FANG, GISP, and GRASPrune, layer-wise allocations remain
identical or highly correlated while the identities of retained FFN channels
change substantially. Thus, similar layer-wise allocation does not uniquely
determine which channels survive. Appendix~\ref{app:credit_diagnostic}
summarizes these comparisons and their protocol scope.

\subsection{Main Results on Qwen-2.5-7B}
\label{subsec:qwen_main}

\begin{table}[H]
\centering
\scriptsize
\setlength{\tabcolsep}{2.8pt}
\caption{Qwen-2.5-7B direct Step-0 mathematical accuracy (\%) using the calibration corpus shown for each method. Values are mean $\pm$ sample
SD over structural seeds 42, 100, and 2026. Calibration data are shown in
parentheses after each method. Task-input calibration is reported at
the 40\% operating point below, with the complete sweep in
Appendix~\ref{tab:app_qwen_taskcal_match}.}
\label{tab:main_qwen_step0}
\begin{tabular*}{\linewidth}{@{\extracolsep{\fill}}clccc@{}}
\toprule
\textbf{Sparsity} & \textbf{Method} & \textbf{GSM8K} & \textbf{AQuA} & \textbf{MathQA} \\
\midrule
\multirow{5}{*}{30\%}
& Global RKU (Wiki16K) & $\mathbf{40.84\pm0.73}$ & $31.10\pm0.39$ & $\mathbf{40.51\pm0.08}$ \\
& GISP (C4-16K) & $26.03\pm13.26$ & $26.38\pm1.18$ & $39.06\pm3.42$ \\
& NIRVANA (BookCorpus) & $7.28\pm0.99$ & $\mathbf{32.55\pm0.99}$ & $30.70\pm0.62$ \\
& Wanda-Struct (Wiki16K) & $9.70\pm0.00$ & $25.59\pm0.00$ & $37.60\pm0.02$ \\
& Taylor-FO (Wiki16K) & $27.02\pm0.66$ & $24.67\pm0.23$ & $36.97\pm0.11$ \\
\midrule
\multirow{5}{*}{40\%}
& Global RKU (Wiki16K) & $\mathbf{12.26\pm0.19}$ & $\mathbf{27.82\pm0.23}$ & $\mathbf{34.66\pm0.26}$ \\
& GISP (C4-16K) & $3.64\pm3.15$ & $25.85\pm2.24$ & $33.90\pm2.33$ \\
& NIRVANA (BookCorpus) & $0.71\pm0.31$ & $21.65\pm2.08$ & $24.98\pm0.28$ \\
& Wanda-Struct (Wiki16K) & $1.97\pm0.00$ & $24.41\pm0.00$ & $31.29\pm0.00$ \\
& Taylor-FO (Wiki16K) & $8.39\pm0.16$ & $24.54\pm0.23$ & $31.90\pm0.12$ \\
\midrule
\multirow{5}{*}{50\%}
& Global RKU (Wiki16K) & $\mathbf{2.30\pm0.18}$ & $\mathbf{26.51\pm0.23}$ & $\mathbf{30.28\pm0.12}$ \\
& GISP (C4-16K) & $0.78\pm0.39$ & $25.20\pm2.19$ & $29.53\pm1.85$ \\
& NIRVANA (BookCorpus) & $0.28\pm0.36$ & $23.88\pm3.35$ & $21.07\pm0.25$ \\
& Wanda-Struct (Wiki16K) & $1.44\pm0.00$ & $21.65\pm0.00$ & $28.04\pm0.00$ \\
& Taylor-FO (Wiki16K) & $1.77\pm0.39$ & $23.49\pm0.23$ & $27.64\pm0.15$ \\
\bottomrule
\end{tabular*}
\end{table}

\paragraph{Step-0 Performance.}
Table~\ref{tab:main_qwen_step0} shows that Global RKU obtains the highest direct accuracy in eight of nine mathematical cells and is higher than GISP in all nine cells of this cross-corpus comparison. Across the nine cells, RKU averages 27.36\% versus 23.37\% for GISP, a 3.99-point gap. Because the calibration corpora differ, we use the task-input comparison in \secref{subsec:cross_family_main} for direct RKU--GISP comparison. The advantage over the highest non-RKU entry in the table is largest on GSM8K at 30\% sparsity (+13.82 points) and narrows to 0.53--1.31 points across the three tasks at 50\%, indicating that the observed advantage is not confined to a single tested sparsity level. RKU is also stable across structural seeds, with a mean SD of 0.27 points and a maximum of 0.73, compared with 3.33 and 13.26 points for GISP. Taylor-FO has similarly small variance. Wanda's near-zero SD has a different origin: under locked Wiki16K calibration, its deterministic score and fixed per-layer quota produce the same retained mask for all three structural seeds; Appendix~\ref{app:qwen_external_results} documents this seed invariance.

\begin{table}[!htbp]
\centering
\scriptsize
\setlength{\tabcolsep}{3pt}
\renewcommand{\arraystretch}{0.92}
\caption{Qwen-2.5-7B paired MathQA OOD recovery after 100/150 LoRA steps.
Each entry averages GSM8K$\rightarrow$MathQA and
AQuA$\rightarrow$MathQA transfer accuracy over seeds 42, 100, and 2026.
GISP uses C4-16K; NIRVANA uses its native BookCorpus protocol.}
\label{tab:main_qwen_external_recovery}
\begin{tabular*}{\linewidth}{@{\extracolsep{\fill}}clcc@{}}
\toprule
\textbf{Sparsity} & \textbf{Method} & \textbf{OOD@100} & \textbf{OOD@150} \\
\midrule
\multirow{5}{*}{30\%}
& Global RKU (Wiki16K) & $\mathbf{40.09\pm0.24}$ & $\mathbf{40.22\pm0.23}$ \\
& GISP (C4-16K) & $39.13\pm2.91$ & $39.06\pm2.89$ \\
& NIRVANA (BookCorpus) & $32.79\pm0.46$ & $32.79\pm0.44$ \\
& Wanda-Struct (Wiki16K) & $39.17\pm0.17$ & $39.46\pm0.28$ \\
& Taylor-FO (Wiki16K) & $38.01\pm0.08$ & $38.26\pm0.02$ \\
\midrule
\multirow{5}{*}{40\%}
& Global RKU (Wiki16K) & $\mathbf{35.50\pm0.25}$ & $\mathbf{35.56\pm0.21}$ \\
& GISP (C4-16K) & $34.04\pm2.58$ & $34.30\pm2.55$ \\
& NIRVANA (BookCorpus) & $27.93\pm0.96$ & $27.95\pm0.94$ \\
& Wanda-Struct (Wiki16K) & $33.59\pm0.22$ & $33.85\pm0.03$ \\
& Taylor-FO (Wiki16K) & $33.16\pm0.08$ & $33.47\pm0.22$ \\
\midrule
\multirow{5}{*}{50\%}
& Global RKU (Wiki16K) & $\mathbf{31.62\pm0.25}$ & $\mathbf{31.68\pm0.20}$ \\
& GISP (C4-16K) & $29.92\pm2.07$ & $30.01\pm2.01$ \\
& NIRVANA (BookCorpus) & $23.86\pm1.30$ & $23.81\pm1.43$ \\
& Wanda-Struct (Wiki16K) & $29.63\pm0.14$ & $29.92\pm0.24$ \\
& Taylor-FO (Wiki16K) & $29.26\pm0.25$ & $29.64\pm0.11$ \\
\bottomrule
\end{tabular*}
\end{table}

\paragraph{Held-Out Recovery.}
Table~\ref{tab:main_qwen_external_recovery} evaluates OOD transfer after separate LoRA adaptation on GSM8K and AQuA, with both recovered models tested on held-out MathQA and averaged by seed. In this cross-protocol comparison, RKU is highest in all six
sparsity--checkpoint cells. Relative to the best competing entry in each cell, its margin is 0.92/0.76 points at 30\%, 1.46/1.26 at 40\%, and 1.70/1.67 at 50\% for 100/150 steps. Its SD remains between 0.20 and 0.25 points. Increasing the recovery budget from 100 to 150 steps changes RKU by only 0.06--0.13 points, so the held-out score has largely plateaued by 100 steps rather than depending on extended adaptation.

\begin{table}[H]
\centering
\scriptsize
\setlength{\tabcolsep}{2.2pt}
\caption{Qwen-2.5-7B in-domain recovery accuracy (\%) after GSM8K or AQuA
LoRA adaptation. Values are mean $\pm$ sample SD over seeds 42, 100, and 2026.
G@100/G@150 denote GSM8K ID accuracy and A@100/A@150 denote AQuA ID accuracy.
NIRVANA retains its native pruning/calibration protocol; GISP uses the protocol
defined in the main evaluation section.}
\label{tab:main_qwen_id_recovery}
\begin{tabular*}{\linewidth}{@{\extracolsep{\fill}}clcccc@{}}
\toprule
\textbf{Sparsity} & \textbf{Method} & \textbf{G@100} & \textbf{A@100} & \textbf{G@150} & \textbf{A@150} \\
\midrule
\multirow{5}{*}{30\%}
& Global RKU (Wiki16K) & $\mathbf{50.24\pm2.05}$ & $\mathbf{30.97\pm0.60}$ & $\mathbf{50.87\pm2.95}$ & $\mathbf{30.18\pm1.14}$ \\
& GISP (C4-16K) & $36.57\pm10.32$ & $25.59\pm1.18$ & $38.51\pm9.94$ & $26.12\pm1.14$ \\
& NIRVANA (BookCorpus) & $20.02\pm1.30$ & $28.08\pm3.18$ & $20.80\pm0.39$ & $28.61\pm3.82$ \\
& Wanda-Struct (Wiki16K) & $35.23\pm1.23$ & $25.07\pm0.60$ & $38.97\pm1.14$ & $25.72\pm0.45$ \\
& Taylor-FO (Wiki16K) & $32.40\pm1.21$ & $27.03\pm0.45$ & $35.61\pm1.30$ & $27.30\pm0.23$ \\
\midrule
\multirow{5}{*}{40\%}
& Global RKU (Wiki16K) & $\mathbf{30.78\pm0.93}$ & $\mathbf{25.33\pm0.45}$ & $\mathbf{32.70\pm0.61}$ & $\mathbf{25.20\pm0.79}$ \\
& GISP (C4-16K) & $16.12\pm7.02$ & $24.15\pm1.98$ & $17.87\pm7.59$ & $24.15\pm0.99$ \\
& NIRVANA (BookCorpus) & $7.71\pm0.69$ & $23.88\pm2.79$ & $8.14\pm0.46$ & $23.88\pm3.39$ \\
& Wanda-Struct (Wiki16K) & $17.61\pm0.37$ & $25.07\pm0.45$ & $21.03\pm0.39$ & $24.54\pm0.60$ \\
& Taylor-FO (Wiki16K) & $15.34\pm0.62$ & $24.15\pm0.60$ & $17.36\pm1.32$ & $24.41\pm0.39$ \\
\midrule
\multirow{5}{*}{50\%}
& Global RKU (Wiki16K) & $\mathbf{15.24\pm2.28}$ & $24.54\pm0.60$ & $\mathbf{16.20\pm1.20}$ & $24.41\pm0.39$ \\
& GISP (C4-16K) & $5.31\pm2.13$ & $23.49\pm0.60$ & $6.12\pm2.87$ & $24.02\pm0.79$ \\
& NIRVANA (BookCorpus) & $2.15\pm0.63$ & $19.95\pm1.59$ & $2.45\pm0.12$ & $19.55\pm4.10$ \\
& Wanda-Struct (Wiki16K) & $7.23\pm0.39$ & $\mathbf{25.59\pm0.39}$ & $10.51\pm0.31$ & $\mathbf{26.12\pm0.23}$ \\
& Taylor-FO (Wiki16K) & $6.44\pm0.67$ & $22.31\pm0.23$ & $9.78\pm0.72$ & $22.83\pm0.00$ \\
\bottomrule
\end{tabular*}
\end{table}

\paragraph{In-Domain Recovery.}
Table~\ref{tab:main_qwen_id_recovery} reports the source-task recovery endpoints. RKU shows a substantially larger response on GSM8K than on
AQuA, while the held-out MathQA behavior is summarized in Table~\ref{tab:main_qwen_external_recovery}.

\subsection{Non-Mathematical Reasoning}
\label{subsec:qwen_breadth_main}

\begin{table}[H]
\centering
\scriptsize
\setlength{\tabcolsep}{5pt}
\caption{Qwen-2.5-7B Step-0 accuracy (\%) on three non-mathematical reasoning
tasks across 30/40/50\% structured sparsity. Values are mean $\pm$ sample SD
over seeds 42, 100, and 2026; no recovery is applied.}
\label{tab:main_qwen_nonmath_50}
\begin{tabular*}{\linewidth}{@{\extracolsep{\fill}}clccc@{}}
\toprule
\textbf{Sparsity} & \textbf{Method} & \textbf{ReClor} &
\textbf{CSQA} & \textbf{LogiQA} \\
\midrule
\multirow{3}{*}{30\%}
& Global RKU & $\mathbf{65.27\pm0.99}$ & $\mathbf{73.96\pm0.14}$ & $\mathbf{47.47\pm0.15}$ \\
& Wanda-Struct & $50.60\pm0.00$ & $69.86\pm0.00$ & $43.78\pm0.00$ \\
& Taylor-FO & $56.33\pm0.42$ & $69.40\pm0.21$ & $46.13\pm0.09$ \\
\midrule
\multirow{3}{*}{40\%}
& Global RKU & $\mathbf{50.87\pm0.70}$ & $62.90\pm0.28$ & $\mathbf{42.19\pm0.49}$ \\
& Wanda-Struct & $47.00\pm0.00$ & $\mathbf{63.39\pm0.00}$ & $38.40\pm0.00$ \\
& Taylor-FO & $43.87\pm0.42$ & $59.30\pm0.14$ & $38.66\pm0.54$ \\
\midrule
\multirow{3}{*}{50\%}
& Global RKU & $\mathbf{45.00\pm0.20}$ & $\mathbf{50.70\pm0.14}$ & $\mathbf{38.15\pm0.44}$ \\
& Wanda-Struct & $38.00\pm0.00$ & $46.52\pm0.00$ & $33.33\pm0.00$ \\
& Taylor-FO & $35.40\pm0.35$ & $47.50\pm0.30$ & $34.56\pm0.15$ \\
\bottomrule
\end{tabular*}
\end{table}

\paragraph{Non-Mathematical Reasoning.}
Table~\ref{tab:main_qwen_nonmath_50} evaluates the same Step-0 topologies on
ReClor, CommonsenseQA, and LogiQA, none used for calibration or recovery.
Among the three matched FFN-only methods in Table~\ref{tab:main_qwen_nonmath_50},
RKU has the highest mean accuracy in eight of nine cells and in all three tasks at
50\% sparsity. These results provide complementary evidence on the tested
non-mathematical reasoning tasks.

\subsection{Calibration and Cross-Family Evaluation}
\label{subsec:cross_family_main}

\paragraph{Questions-Only Calibration.}
We compare RKU and GISP using the same GSM8K-Q16K calibration, constructed
from the GSM8K training split's \texttt{question} field only, with no answers or
solution traces. At the 40\% operating point, RKU reaches $45.96\pm1.55$
Mean3 versus $42.17\pm1.27$ for GISP, a 3.79-point margin. Under Full16K
calibration with full solutions, GISP is higher across all three sparsity levels by
0.37--2.63 points. This contrast shows that calibration supervision can change
the relative ordering of the two methods. The complete 30--50\% comparison is
reported in Appendix~\ref{tab:app_qwen_taskcal_match}.

\begin{table}[H]
\centering
\scriptsize
\setlength{\tabcolsep}{2.0pt}
\caption{Label-free calibration transfer for Global RKU. Each cell reports
Wiki16K $\rightarrow$ GSM8K-Q16K Mean3 (mean $\pm$ sample SD over seeds 42,
100, and 2026), followed by the Q16K gain in parentheses. Q16K is built
from the GSM8K training split using only the \texttt{question} field; no answers
or solution traces are included.}
\label{tab:main_rku_label_free_transfer}
\resizebox{\linewidth}{!}{%
\begin{tabular}{c|c|c|c}
\toprule
\textbf{Sparsity} & \textbf{Qwen-2.5-7B} & \textbf{Llama-3.1-8B} & \textbf{Gemma-2-9B} \\
\midrule
30\% & $37.49\!\pm\!0.14 \rightarrow \mathbf{47.66\!\pm\!0.57}$ $(+10.17)$
     & $18.40\!\pm\!0.18 \rightarrow \mathbf{20.67\!\pm\!0.24}$ $(+2.27)$
     & $31.38\!\pm\!0.73 \rightarrow \mathbf{35.93\!\pm\!0.28}$ $(+4.55)$ \\
40\% & $24.91\!\pm\!0.08 \rightarrow \mathbf{45.96\!\pm\!1.55}$ $(+21.05)$
     & $16.74\!\pm\!0.15 \rightarrow \mathbf{18.13\!\pm\!0.70}$ $(+1.39)$
     & $25.57\!\pm\!0.65 \rightarrow \mathbf{29.03\!\pm\!0.57}$ $(+3.46)$ \\
50\% & $19.70\!\pm\!0.04 \rightarrow \mathbf{39.35\!\pm\!1.21}$ $(+19.65)$
     & $15.84\!\pm\!0.13 \rightarrow \mathbf{16.33\!\pm\!0.23}$ $(+0.49)$
     & $19.69\!\pm\!0.26 \rightarrow \mathbf{23.03\!\pm\!0.43}$ $(+3.34)$ \\
\bottomrule
\end{tabular}}
\end{table}

\paragraph{Cross-Family Calibration Transfer.}
Table~\ref{tab:main_rku_label_free_transfer} provides the replicated test corresponding to the mechanism probe above without answers, solution traces, or target-loss labels. Questions-only Q16K, constructed from the GSM8K training split using only the \texttt{question} field, improves RKU over Wiki16K in all nine model--sparsity cells, with average gains of +16.96 points on Qwen, +3.78 on Gemma, and +1.38 on Llama. Task-level effects are reported in Tables~\ref{tab:app_llama_rku_taskinput_extension} and \ref{tab:app_gemma_step0_complete}; the aggregate result supports label-free calibration-distribution sensitivity across all three backbones \cite{ji2025calibration}.

\paragraph{Physical Deployment.} Detailed physical parameter, memory, time-to-first-token (TTFT), and decode-throughput measurements are reported in Appendix~\ref{app:resource_accounting}.

\subsection{Relative Normalization Ablation}
\label{subsec:signal_factorial_main}

\begin{table}[H]
\centering
\scriptsize
\setlength{\tabcolsep}{2.0pt}
\caption{Qwen-2.5-7B signal-factorial ablation. Mean3 averages GSM8K, AQuA,
and MathQA within seed, then across seeds 42, 100, and 2026. Final-hidden-state
rows form a $2\times2$ factorial over $w_{l,c}$ and $Z_l^{-1}$; LM-CE changes
only the probe. Calibration master, FFN-only pruning space, global budget,
selection, seeds, and evaluator are fixed. $\Delta_{\mathrm{rel}}$ is RKU minus
$w_{l,c}\mathcal A_{l,c}$ (RKU w/o Relative Calibration).}
\label{tab:main_signal_factorial}
\begin{tabular*}{\linewidth}{@{\extracolsep{\fill}}llccc@{}}
\toprule
\textbf{Variant} & \textbf{Probe / score} & \textbf{40\%} & \textbf{50\%} & \textbf{Change} \\
\midrule
$\mathcal A_{l,c}$ & final-hidden-state $\mathcal A_{l,c}$ & $17.59\pm1.20$ & $16.75\pm0.45$ & no $w_{l,c}$, no $Z_l$ \\
$w_{l,c}\mathcal A_{l,c}$ & final-hidden-state $w_{l,c}\mathcal A_{l,c}$ & $17.94\pm1.67$ & $16.80\pm0.34$ & no $Z_l$ \\
$\mathcal A_{l,c}/Z_l$ & final-hidden-state $\mathcal A_{l,c}/Z_l$ & $27.19\pm5.40$ & $20.58\pm2.89$ & no $w_{l,c}$ \\
LM-CE & LM-CE $w_{l,c}\mathcal A_{l,c}/Z_l$ & $28.40\pm3.36$ & $21.03\pm2.35$ & replace probe \\
\textbf{RKU} & final-hidden-state $w_{l,c}\mathcal A_{l,c}/Z_l$ &
$\mathbf{32.36\pm3.44}$ & $\mathbf{22.33\pm2.82}$ & complete score \\
\midrule
$\boldsymbol{\Delta}_{\mathrm{rel}}$ & RKU $-$ $w_{l,c}\mathcal A_{l,c}$ & $\mathbf{+14.42}$ & $\mathbf{+5.53}$ & relative calibration \\
\bottomrule
\end{tabular*}
\end{table}

\begin{figure}[H]
\centering
\noindent\begin{minipage}[t]{0.49\linewidth}
\centering
\captionsetup{font=small,skip=3pt}
\captionof{table}{Qwen-2.5-7B common-seed Step-0 margins: Global RKU minus
RKU w/o Relative. All 27 seed--task margins are positive.}
\label{tab:main_global_step0_seed_deltas}
\vspace{-2pt}
\fontsize{7.4}{7.9}\selectfont
\setlength{\tabcolsep}{1.0pt}
\renewcommand{\arraystretch}{0.98}
\begin{tabular*}{\linewidth}{@{\extracolsep{\fill}}ccccc@{}}
\toprule
\textbf{Sparsity} & \textbf{Seed} & $\boldsymbol{\Delta}$ \textbf{GSM8K}
& $\boldsymbol{\Delta}$ \textbf{AQuA} & $\boldsymbol{\Delta}$ \textbf{MathQA} \\
\midrule
30\% & 42   & +20.32 & +4.33 & +3.05 \\
30\% & 100  & +20.85 & +3.15 & +2.88 \\
30\% & 2026 & +20.47 & +3.54 & +2.68 \\
\midrule
40\% & 42   & +11.60 & +3.15 & +5.90 \\
40\% & 100  & +11.45 & +2.76 & +6.37 \\
40\% & 2026 & +11.75 & +3.54 & +5.43 \\
\midrule
50\% & 42   & +1.90 & +3.15 & +4.96 \\
50\% & 100  & +2.20 & +2.76 & +5.09 \\
50\% & 2026 & +1.82 & +1.97 & +4.89 \\
\bottomrule
\end{tabular*}
\end{minipage}\hfill
\begin{minipage}[t]{0.49\linewidth}
\centering
\vspace{0.72cm}
\includegraphics[width=\linewidth]{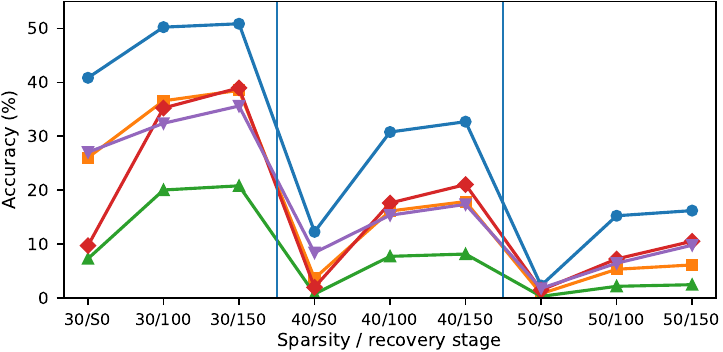}
\vspace{-3pt}
{\footnotesize\textbf{(a)} GSM8K ID recovery\par}
\end{minipage}

\vspace{2pt}
\noindent\begin{minipage}[t]{0.49\linewidth}
\centering
\includegraphics[width=\linewidth]{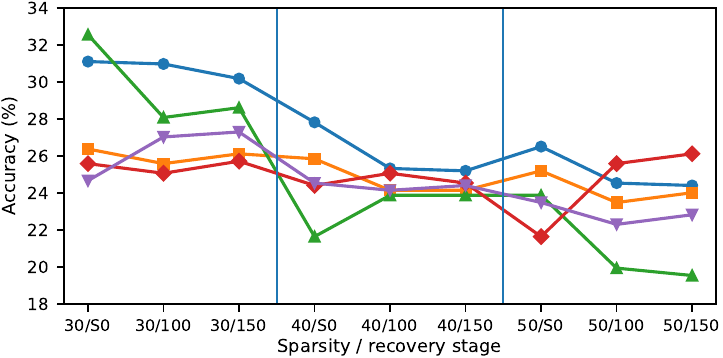}
\vspace{-3pt}
{\footnotesize\textbf{(b)} AQuA ID recovery\par}
\end{minipage}\hfill
\begin{minipage}[t]{0.49\linewidth}
\centering
\includegraphics[width=\linewidth]{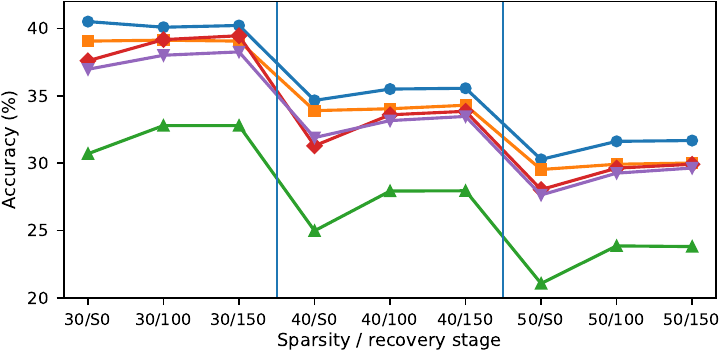}
\vspace{-3pt}
{\footnotesize\textbf{(c)} Paired MathQA OOD recovery\par}
\end{minipage}

\vspace{1pt}
\includegraphics[width=0.90\linewidth]{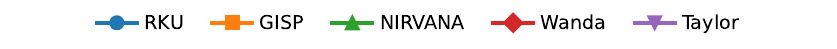}
\vspace{-3pt}
\captionsetup{type=figure,font=small,skip=3pt}
\caption{Qwen-2.5-7B recovery trajectories: (a) GSM8K in-domain recovery,
(b) AQuA in-domain recovery, and (c) paired MathQA OOD recovery. For (c),
OOD@100/150 averages GSM8K$\rightarrow$MathQA and AQuA$\rightarrow$MathQA transfer.}
\label{fig:qwen_recovery_grid}
\end{figure}

\paragraph{Relative Calibration Effect.}
At 40/50\% sparsity, adding $Z_l^{-1}$ to $\mathcal A_{l,c}$ gives
+9.60/+3.83 Mean3 points, while adding $w_{l,c}$ gives +0.35/+0.05. The
interaction is non-additive: $w_{l,c}$ contributes +5.17/+1.75 after
normalization, whereas relative calibration (RKU $-$ $w_{l,c}\mathcal A_{l,c}$)
contributes +14.42/+5.53 after weighting, the larger paired contrast at both
sparsities. The final-hidden-state probe adds +3.96/+1.30 over LM-CE. All 27
common-seed task margins are positive (Table~\ref{tab:main_global_step0_seed_deltas}).

\section{Conclusion}
\label{sec:conclusion}

Global structured FFN pruning needs retention credit for channels competing across
layers under one budget. Global RKU uses final-hidden-state activation--gradient
participation from unlabeled calibration text and mitigates block-common scale before
model-wide selection. On the evaluated Qwen-2.5-7B protocols, it obtains the reported
gains in mathematical and non-mathematical reasoning, held-out transfer, low-budget
recovery, and physical deployment. Matched ablations identify relative normalization as
a substantial contributor on the primary mathematical suite; broader reasoning and
cross-family results show that its effect varies with sparsity and architecture, supporting
functional, relative credit in the evaluated global structured FFN-pruning setting.

\section*{Reproducibility Statement}

\Secref{sec:methodology} and \algref{alg:global_rku} specify the scoring rule,
relative calibration, and mask construction. Appendix~\ref{app:supplementary_results}
records calibration construction, seed semantics, structured-control definitions,
evaluator settings, and full recovery hyperparameters. Appendix~\ref{app:resource_accounting}
specifies physicalization and hardware measurement settings; Appendices~\ref{app:qwen_external_results}--\ref{app:llama_complete_results}
record external-comparison provenance and cross-family diagnostics.

\section*{AI Use Statement}

Generative AI tools were used for literature search and related-work discovery,
reference checking, experimental implementation including code generation, and
drafting and editing parts of the manuscript. They were not used to generate data
or experimental results, and no other required-disclosure research tasks involved
generative AI. The authors reviewed all AI-assisted text, code, references, and
analyses and take responsibility for the final paper.

\clearpage

\bibliographystyle{iclr2027_conference}
\bibliography{rku_references_complete}

\clearpage
\appendix

\section{Additional Methodological Details}
\label{app:method_derivations}

\subsection{First-Order Motivation for Directional Participation}

AGF assigns utility to a candidate feature through its directional contribution
to the training objective \citep{kunin2025alternating}. Let
$f_{\mathcal S_{\mathrm{act}}}$ denote the predictor formed by the currently active feature
set $\mathcal S_{\mathrm{act}}$, let $f_i$ be a candidate feature, and let
$r=y-f_{\mathcal S_{\mathrm{act}}}(x)$ be the residual. Under the squared loss
$\Ls(f)=\frac{1}{2}\E_{(x,y)}\|y-f(x)\|_2^2$, its utility is
\begin{equation}
\mathcal U_i
=
\E_{(x,y)}\langle f_i(x),r\rangle
=
-\left.
\partial_\lambda
\Ls(f_{\mathcal S_{\mathrm{act}}}+\lambda f_i)
\right|_{\lambda=0},
\label{eq:agf_utility_prelim}
\end{equation}
where $\lambda$ controls the contribution of the candidate feature.

\subsection{Approximate Block-Scale Factorization}

To make the approximation above explicit, write
$\mathcal A_{l,c}=\alpha_l q_{l,c}(1+\xi_{l,c})$ with
$\alpha_l>0$, $q_{l,c}\ge0$, $Q_l=\sum_j q_{l,j}>0$, and
$|\xi_{l,c}|\le\varepsilon_l<1$. Let
$\bar\xi_l=Q_l^{-1}\sum_j q_{l,j}\xi_{l,j}$.
When the numerical floor does not alter the denominator, i.e.,
\[
\alpha_l Q_l(1+\bar\xi_l)\ge10^{-12},
\]
we have
\[
Z_l=\alpha_lQ_l(1+\bar\xi_l),
\]
and therefore
\begin{equation}
\frac{\mathcal A_{l,c}}{Z_l}
=
\frac{q_{l,c}}{Q_l}
\frac{1+\xi_{l,c}}{1+\bar\xi_l}.
\label{eq:relative_credit_factorization}
\end{equation}
Consequently,
\begin{equation}
\left|
\frac{\mathcal A_{l,c}/Z_l}{q_{l,c}/Q_l}-1
\right|
=
\left|
\frac{\xi_{l,c}-\bar\xi_l}{1+\bar\xi_l}
\right|
\le
\frac{2\varepsilon_l}{1-\varepsilon_l}.
\label{eq:relative_credit_bound}
\end{equation}
Thus an exact block-common multiplier cancels in the nondegenerate regime,
while bounded channel-wise deviations perturb the normalized score by the
amount above. If $\alpha_l Q_l(1+\bar\xi_l)<10^{-12}$, then
$Z_l=10^{-12}$ by construction; this branch serves only as a safeguard
against division by a near-zero block aggregate, and the block-scale
cancellation interpretation is not invoked. Here $q_{l,c}$ denotes the
channel-specific component of the factorization.
\Secref{subsec:signal_factorial_main} compares the normalized score with
its unnormalized counterpart.

\section{Experimental Details and Relative-Credit Mechanism Audit}

\label{app:supplementary_results}

\paragraph{Appendix Organization.}
The supporting experiments are organized by the question they answer rather than by execution chronology. We first record resource and implementation scope, then isolate the normalization factor under a matched control, expose seed-level results, and finally examine recovery and cross-family behavior. This ordering distinguishes mechanism comparisons from protocol-conditioned external references.

\subsection{Physical Hardware and Pareto Closure}
\label{app:resource_accounting}

\paragraph{Physicalization Protocol.}
The hardware closure measures exported dense submodels rather than timing zero-masked dense tensors. For RKU, Wanda-Struct, and Taylor-FO, a retained FFN index set slices the corresponding rows of the gate/up projections and the same columns of the down projection. For GISP and NIRVANA, the native reproduced masks contain both FFN channels and Qwen GQA groups. Each retained KV group keeps its associated query-head group; the exporter slices the corresponding rows of $\mW_Q,\mW_K,\mW_V$, the matching columns of $\mW_O$, and the retained FFN rows/columns. Consequently, the reported parameter counts reflect the realized physical shape and need not equal the nominal pruning ratio. All timing uses a single NVIDIA GeForce RTX 4090, BF16 weights, batch size 1, PyTorch SDPA, 128 cached decode tokens, and prompt lengths 512 or 2048. TTFT is the synchronized full-prompt forward pass that produces first-token logits and KV cache; prefill throughput is prompt tokens divided by this latency. Decode throughput times only the subsequent cached 128-token autoregressive loop. Peak allocated memory includes the prompt and the complete 128-token decode cache.

\begin{table}[H]
\centering
\appendixtablestyle
\caption{Physical deployment protocol used for the Qwen-2.5-7B hardware/Pareto closure.}
\label{tab:app_physical_protocol}
\begin{tabularx}{\linewidth}{@{}lY@{}}
\toprule
\textbf{Item} & \textbf{Setting} \\
\midrule
GPU & one NVIDIA GeForce RTX 4090 (24 GB) \\
Precision & BF16 \\
Batch size & 1 \\
Attention kernel & PyTorch SDPA \\
Prompt length & 512 and 2048 tokens \\
Decode length & 128 cached autoregressive tokens \\
Peak-memory window & full prompt plus all 128 decode tokens \\
RKU/Wanda/Taylor dynamic seeds & 42, 100, 2026 \\
GISP/NIRVANA dynamic seed & 42 \\
RKU/Wanda/Taylor physical space & coupled SwiGLU FFN channels \\
GISP/NIRVANA physical space & retained GQA groups plus retained FFN channels \\
\bottomrule
\end{tabularx}
\end{table}

\begin{table}[H]
\centering
\scriptsize
\setlength{\tabcolsep}{2.0pt}
\caption{Qwen-2.5-7B physical deployment at the 40\% operating point. Mean3
averages GSM8K, AQuA, and MathQA over structural seeds 42, 100, and 2026;
hardware columns use the seed-42 topology for a method-to-method
comparison. Peak memory includes the complete 128-token decode cache. The
512/2048 entries correspond to prompt length. The Dense row is the unpruned
Qwen-2.5-7B reference measured under the same hardware protocol.}
\label{tab:app_qwen_physical_40}
\begin{tabular*}{\linewidth}{@{\extracolsep{\fill}}lrrrrr@{}}
\toprule
\textbf{Method} & \textbf{Mean3} & \textbf{Params(B)} & \textbf{Peak GB} &
\textbf{TTFT ms} & \textbf{Decode tok/s} \\
\midrule
Dense & \resna & 7.616 & 14.33/14.62 & 53.41/190.60 & 56.19/55.26 \\
Global RKU & \textbf{24.91} & 5.334 & 10.02/10.25 & 41.69/150.68 & \textbf{69.49/69.73} \\
GISP & 21.13 & 5.520 & 10.40/10.68 & 43.57/156.18 & 67.51/67.89 \\
NIRVANA & 15.78 & \textbf{5.005} & \textbf{9.42/9.66} & \textbf{35.40/123.78} & 68.77/69.72 \\
Wanda-Struct & 19.22 & 5.334 & 10.02/10.25 & 35.92/134.82 & 68.88/68.60 \\
Taylor-FO & 21.61 & 5.334 & 10.02/10.25 & 35.97/134.84 & 68.78/69.19 \\
\bottomrule
\end{tabular*}
\end{table}

\paragraph{Quality--Kernel Tradeoff.}
Table~\ref{tab:app_qwen_physical_40} closes the structural results with actual dense-shape execution rather than zero-mask timing. At 40\% nominal sparsity, RKU exceeds GISP by 3.78 Mean3 points in this
cross-protocol deployment comparison while using fewer physical parameters (5.334B vs. 5.520B), less peak memory at both prompt lengths, lower TTFT (41.69 vs. 43.57 ms at 512 and 150.68 vs. 156.18 ms at 2048), and higher decode throughput. The comparison to fixed-quota FFN baselines is different: Wanda and Taylor have essentially the same parameter count and memory as RKU but lower TTFT because their retained widths are regular across layers, whereas RKU spends the same global budget nonuniformly. NIRVANA forms a more aggressive efficiency endpoint: it is smaller and faster but loses 9.13 Mean3 points relative to RKU at 40\%. The result is a quality--kernel tradeoff: cross-layer credit allocation improves retained quality, while the resulting irregular layer widths need not maximize dense-kernel efficiency.

\paragraph{Effective Parameter Reduction.}
Because the methods prune different structural spaces, nominal 30/40/50\% labels are not interchangeable with total-parameter reduction. RKU physically contains 5.905/5.334/4.764B parameters, corresponding to 22.47/29.96/37.44\% reduction from the 7.616B dense model. At the same nominal labels, GISP contains 6.051/5.520/4.981B parameters (20.54/27.52/34.59\% reduction), while NIRVANA contains 5.656/5.005/4.355B (25.73/34.27/42.82\%). 

\begin{table}[H]
\centering
\scriptsize
\setlength{\tabcolsep}{2.2pt}
\caption{Physical Qwen-2.5-7B results with a 512-token prompt. Quality is the
three-task Step-0 Mean3 from Table~\ref{tab:main_qwen_step0}. RKU, Wanda, and
Taylor hardware values are mean $\pm$ sample SD over structural seeds
42/100/2026; GISP and NIRVANA use the seed-42 physical export.}
\label{tab:app_physical_512}
\begin{tabular*}{\linewidth}{@{\extracolsep{\fill}}clrrrrr@{}}
\toprule
\textbf{Sp.} & \textbf{Method} & \textbf{Mean3} & \textbf{Params(B)} & \textbf{Peak GB} & \textbf{TTFT ms} & \textbf{Decode tok/s} \\
\midrule
\multirow{5}{*}{30\%}
& Global RKU & \textbf{37.48} & 5.905 & $11.09\pm0.00$ & $46.58\pm1.02$ & $66.70\pm0.10$ \\
& GISP & 30.49 & 6.051 & 11.38 & 45.16 & 66.12 \\
& NIRVANA & 23.51 & 5.656 & 10.64 & 38.29 & 66.44 \\
& Wanda-Struct & 24.30 & 5.905 & $11.09\pm0.00$ & $51.36\pm0.03$ & $68.67\pm0.31$ \\
& Taylor-FO & 29.55 & 5.905 & $11.09\pm0.00$ & $51.39\pm0.02$ & $68.68\pm0.23$ \\
\midrule
\multirow{5}{*}{40\%}
& Global RKU & \textbf{24.91} & 5.334 & $10.02\pm0.00$ & $41.11\pm1.54$ & $69.31\pm0.17$ \\
& GISP & 21.13 & 5.520 & 10.40 & 43.57 & 67.51 \\
& NIRVANA & 15.78 & 5.005 & 9.42 & 35.40 & 68.77 \\
& Wanda-Struct & 19.22 & 5.334 & $10.02\pm0.00$ & $36.03\pm0.10$ & $68.62\pm0.26$ \\
& Taylor-FO & 21.61 & 5.334 & $10.02\pm0.00$ & $36.05\pm0.07$ & $68.61\pm0.15$ \\
\midrule
\multirow{5}{*}{50\%}
& Global RKU & \textbf{19.70} & 4.764 & $8.95\pm0.00$ & $37.55\pm0.26$ & $69.13\pm0.10$ \\
& GISP & 18.50 & 4.981 & 9.40 & 36.73 & 67.41 \\
& NIRVANA & 15.08 & 4.355 & 8.21 & 30.64 & 68.79 \\
& Wanda-Struct & 17.04 & 4.764 & $8.97\pm0.00$ & $31.48\pm0.02$ & $68.80\pm0.19$ \\
& Taylor-FO & 17.63 & 4.764 & $8.97\pm0.00$ & $31.47\pm0.02$ & $68.85\pm0.09$ \\
\bottomrule
\end{tabular*}
\end{table}

\begin{table}[H]
\centering
\scriptsize
\setlength{\tabcolsep}{2.2pt}
\caption{Physical Qwen-2.5-7B results with a 2048-token prompt, under the same
protocol as Table~\ref{tab:app_physical_512}. Peak memory includes the full
128-token decode cache.}
\label{tab:app_physical_2048}
\begin{tabular*}{\linewidth}{@{\extracolsep{\fill}}clrrrrr@{}}
\toprule
\textbf{Sp.} & \textbf{Method} & \textbf{Mean3} & \textbf{Params(B)} & \textbf{Peak GB} & \textbf{TTFT ms} & \textbf{Decode tok/s} \\
\midrule
\multirow{5}{*}{30\%}
& Global RKU & \textbf{37.48} & 5.905 & $11.33\pm0.00$ & $166.65\pm3.61$ & $65.86\pm0.06$ \\
& GISP & 30.49 & 6.051 & 11.67 & 160.75 & 65.05 \\
& NIRVANA & 23.51 & 5.656 & 10.89 & 138.22 & 65.39 \\
& Wanda-Struct & 24.30 & 5.905 & $11.33\pm0.00$ & $181.73\pm0.32$ & $68.66\pm0.03$ \\
& Taylor-FO & 29.55 & 5.905 & $11.33\pm0.00$ & $181.58\pm0.02$ & $68.65\pm0.03$ \\
\midrule
\multirow{5}{*}{40\%}
& Global RKU & \textbf{24.91} & 5.334 & $10.25\pm0.00$ & $148.56\pm5.42$ & $69.78\pm0.05$ \\
& GISP & 21.13 & 5.520 & 10.68 & 156.18 & 67.89 \\
& NIRVANA & 15.78 & 5.005 & 9.66 & 123.78 & 69.72 \\
& Wanda-Struct & 19.22 & 5.334 & $10.25\pm0.00$ & $134.97\pm0.12$ & $68.98\pm0.37$ \\
& Taylor-FO & 21.61 & 5.334 & $10.25\pm0.00$ & $134.90\pm0.05$ & $69.08\pm0.28$ \\
\midrule
\multirow{5}{*}{50\%}
& Global RKU & \textbf{19.70} & 4.764 & $9.15\pm0.00$ & $133.31\pm1.66$ & $69.70\pm0.03$ \\
& GISP & 18.50 & 4.981 & 9.68 & 129.49 & 67.80 \\
& NIRVANA & 15.08 & 4.355 & 8.43 & 106.92 & 69.53 \\
& Wanda-Struct & 17.04 & 4.764 & $9.18\pm0.00$ & $117.00\pm0.01$ & $69.39\pm0.17$ \\
& Taylor-FO & 17.63 & 4.764 & $9.18\pm0.00$ & $117.00\pm0.00$ & $69.45\pm0.18$ \\
\bottomrule
\end{tabular*}
\end{table}

\paragraph{Long-Prompt Deployment.}
The 2048-token experiment does not reverse the 512-token conclusions. RKU's TTFT speedup over dense changes from 1.15/1.30/1.42$\times$ at 512 tokens to 1.14/1.28/1.43$\times$ at 2048 tokens for 30/40/50\% sparsity, with decode throughput remaining approximately 1.19--1.26$\times$ dense. At 40\%, RKU remains faster than GISP at both prompt lengths while retaining higher quality and fewer parameters. NIRVANA is consistently the most compressed of the compared physical models at 40--50\% and therefore reaches the lowest TTFT, but at substantially lower Mean3. Wanda and Taylor obtain lower TTFT than RKU at 40--50\% despite nearly identical parameter counts, showing that realized dense-kernel efficiency depends on the layer-width profile, not only on total parameters.

\paragraph{Physical Equivalence Audit.}
\label{app:physical_equivalence}
For the 15 Qwen method--sparsity exports, we checked whether re-padding each
compact model into the original coordinates reconstructed the corresponding
zero-masked weights; all 15 recorded checks passed exactly. Small BF16 logit differences remain because physical compaction
changes GEMM shapes and accumulation order. We therefore use exact structural
reconstruction, rather than bitwise output equality, as the export correctness
criterion.

\subsection{Experimental Protocol}
\label{app:pruning_scope}

\begin{table}[H]
\centering
\appendixtablestyle
\caption{Unified pruning scope, implementation, recovery, and hardware settings.}
\label{tab:app_pruning_scope}
\begin{tabularx}{\linewidth}{@{}lY@{}}
\toprule
\textbf{Item} & \textbf{Setting} \\
\midrule
Models & Qwen-2.5-7B / Llama-3.1-8B / Gemma-2-9B \\
RKU/Wanda/Taylor pruned modules & gate projection, up projection, and down projection \\
RKU/Wanda/Taylor pruning unit & Coupled SwiGLU intermediate channel \\
GISP/NIRVANA native structures & Coupled SwiGLU channels plus GQA-compatible attention groups \\
Attention in matched RKU controls & Not pruned \\
Depth & All 28 Qwen blocks / all 32 Llama blocks / all 42 Gemma blocks \\
Global selection & Fixed model-wide budget with a 10\% minimum retained width per block in matched RKU controls \\
Implementation & Qwen mathematical/recovery and Llama/Gemma cross-family experiments: structured zero masks; Qwen breadth and physical deployment: physical FFN compaction \\
Sparsity levels & 30\%, 40\%, 50\% FFN channel sparsity for RKU/Wanda/Taylor; native structural budget for GISP/NIRVANA \\
Structural seeds & 42, 100, 2026 \\
Structural RNG & Python, NumPy, PyTorch, and CUDA seeded consistently; deterministic cuDNN execution \\
Recovery & LoRA rank 8, $\alpha=16$, dropout 0, \texttt{q\_proj}/\texttt{v\_proj}; learning rate $2\times10^{-4}$; BF16; micro-batch 1; gradient accumulation 4; max length 512; checkpoints 100/150 \\
Qwen reasoning tasks & ReClor, CommonsenseQA, LogiQA \\
Calibration input & 32 sequences $\times$ 512 tokens $=16{,}384$ tokens; corpus/view reported per table \\
GSM8K-Q16K source & GSM8K training split, \texttt{question} field only; no answers or solution traces \\
Evaluation & lm-evaluation-harness task implementations; GSM8K/AQuA/MathQA 3-shot, non-mathematical reasoning suite 0-shot; evaluation batch size 1 \\
Hardware & Primary Qwen accuracy/recovery: 2$\times$ NVIDIA GeForce RTX 4090 (24 GB each); Qwen physical closure: one RTX 4090; Llama/Gemma cross-family runs: NVIDIA L20 GPUs \\
\bottomrule
\end{tabularx}
\end{table}

\paragraph{Matched Structural Grid.}
Table~\ref{tab:app_pruning_scope} keeps attention unpruned for the matched RKU
mechanism controls and applies each retained FFN channel index jointly to the
gate, up, and down projections over all Transformer blocks. The Qwen runs share
the same sparsity, seed, calibration-budget, and recovery grid; Llama and Gemma
use the corresponding all-block FFN scope for cross-family evaluation. External
GISP and NIRVANA rows retain the protocol differences stated in their tables.

\paragraph{Structural-Seed Semantics.}
The implementation seeds Python, NumPy, PyTorch, and CUDA for each structural
replicate. In the matched Wiki16K mechanism audits, a fixed partition seed
creates one deterministic permutation of the calibration master and assigns
three non-overlapping 32$\times$512-token views to seeds 42, 100, and 2026.
The same seed-specific view is reused across the score variants compared within
that audit.

\paragraph{Structured Local Controls.}
Wanda-Struct is a fixed-per-layer coupled-SwiGLU adaptation of Wanda: for each
block it multiplies the $\ell_2$ norm of a down-projection weight column by the
accumulated $\ell_2$ norm of the corresponding down-projection input activation,
then retains a uniform local top-$k$ quota. Taylor-FO is the matched first-order
control: only down-projection weights receive gradients, the probe is the
final-hidden-state squared norm, and channel saliency sums
$|W_{\mathrm{down}}\odot\nabla W_{\mathrm{down}}|$ over the output dimension.
Both controls apply each retained intermediate index jointly to the gate, up,
and down projections and use fixed per-layer sparsity rather than model-wide
allocation.

\subsection{RKU w/o Relative Calibration Control}
\label{app:relative_calibration_control}

\paragraph{Normalization Control.}
RKU w/o Relative Calibration uses the same pruning unit, participation signal $\mathcal A_{l,c}$, calibration data, model-wide channel budget, minimum retained-width constraint, and selection protocol as Global RKU. Its score is
\begin{equation}
\mathcal{S}^{\mathrm{unnorm}}_{l,c}
=
w_{l,c}\mathcal A_{l,c}
\label{eq:app_unnormalized_score}
\end{equation}
Global RKU differs only by dividing $\mathcal A_{l,c}$ by $Z_l$ before model-wide ranking. The resulting comparison is therefore the direct test of block-relative normalization used throughout the mechanism audit.

\subsection{Global Recovery Diagnostics}

\begin{table}[p]
\centering
\appendixtablestyle
\setlength{\tabcolsep}{1.8pt}
\caption{Complete seed-level Qwen recovery audit for Global RKU and RKU w/o
Relative Calibration. G-ID/A-ID are in-domain GSM8K/AQuA accuracies;
G$\rightarrow$M and A$\rightarrow$M are MathQA OOD accuracies after the two
separate LoRA adaptations; Pair is their within-seed average. No seed-level
rows are omitted.}
\label{tab:app_global_recovery_complete}
\begin{tabular*}{\linewidth}{@{\extracolsep{\fill}}llccrrrrr@{}}
\toprule
\textbf{Method} & \textbf{Sp.} & \textbf{Seed} & \textbf{Step} &
\textbf{G-ID} & \textbf{G$\rightarrow$M} & \textbf{A-ID} &
\textbf{A$\rightarrow$M} & \textbf{Pair} \\
\midrule
\multirow{18}{*}{RKU} & \multirow{6}{*}{30\%} & 42 & 100 & 47.915087 & 39.430486 & 30.314961 & 40.201005 & 39.815746 \\
 &  & 42 & 150 & 47.611827 & 40.100503 & 29.527559 & 40.067002 & 40.083753 \\
 &  & 100 & 100 & 51.781653 & 40.067002 & 31.102362 & 40.335008 & 40.201005 \\
 &  & 100 & 150 & 53.373768 & 39.899497 & 31.496063 & 40.268007 & 40.083752 \\
 &  & 2026 & 100 & 51.023503 & 40.167504 & 31.496063 & 40.335008 & 40.251256 \\
 &  & 2026 & 150 & 51.630023 & 40.603015 & 29.527559 & 40.368509 & 40.485762 \\
\cmidrule(lr){2-9}
 & \multirow{6}{*}{40\%} & 42 & 100 & 31.463230 & 35.946399 & 25.590551 & 35.611390 & 35.778894 \\
 &  & 42 & 150 & 32.069750 & 36.448911 & 25.984252 & 35.142379 & 35.795645 \\
 &  & 100 & 100 & 31.159970 & 35.912898 & 24.803150 & 34.673367 & 35.293132 \\
 &  & 100 & 150 & 33.282790 & 35.845896 & 24.409449 & 34.974874 & 35.410385 \\
 &  & 2026 & 100 & 29.719484 & 36.247906 & 25.590551 & 34.606365 & 35.427135 \\
 &  & 2026 & 150 & 32.752085 & 36.180905 & 25.196850 & 34.773869 & 35.477387 \\
\cmidrule(lr){2-9}
 & \multirow{6}{*}{50\%} & 42 & 100 & 14.404852 & 31.959799 & 24.015748 & 30.720268 & 31.340034 \\
 &  & 42 & 150 & 15.163002 & 32.060302 & 24.015748 & 30.820771 & 31.440537 \\
 &  & 100 & 100 & 17.816528 & 32.696817 & 25.196850 & 30.820771 & 31.758794 \\
 &  & 100 & 150 & 17.513268 & 32.696817 & 24.803150 & 30.887772 & 31.792295 \\
 &  & 2026 & 100 & 13.495072 & 32.160804 & 24.409449 & 31.390285 & 31.775544 \\
 &  & 2026 & 150 & 15.921152 & 32.160804 & 24.409449 & 31.423786 & 31.792295 \\
\midrule
\multirow{18}{*}{w/o Rel.} & \multirow{6}{*}{30\%} & 42 & 100 & 42.228961 & 38.391960 & 27.952756 & 37.219430 & 37.805695 \\
 &  & 42 & 150 & 43.517817 & 38.592965 & 27.952756 & 37.252931 & 37.922948 \\
 &  & 100 & 100 & 43.442002 & 38.793970 & 27.165354 & 36.716918 & 37.755444 \\
 &  & 100 & 150 & 44.124337 & 38.793970 & 27.165354 & 36.984925 & 37.889448 \\
 &  & 2026 & 100 & 43.138741 & 38.760469 & 27.559055 & 37.386935 & 38.073702 \\
 &  & 2026 & 150 & 44.200152 & 38.358459 & 27.165354 & 36.884422 & 37.621441 \\
\cmidrule(lr){2-9}
 & \multirow{6}{*}{40\%} & 42 & 100 & 8.718726 & 28.911223 & 23.622047 & 27.470687 & 28.190955 \\
 &  & 42 & 150 & 8.491281 & 29.212730 & 24.015748 & 27.537688 & 28.375209 \\
 &  & 100 & 100 & 8.794541 & 29.581240 & 24.015748 & 27.906198 & 28.743719 \\
 &  & 100 & 150 & 8.491281 & 29.346734 & 24.015748 & 28.040201 & 28.693468 \\
 &  & 2026 & 100 & 8.491281 & 29.614740 & 23.622047 & 27.973199 & 28.793970 \\
 &  & 2026 & 150 & 8.188021 & 29.782245 & 24.015748 & 27.604690 & 28.693468 \\
\cmidrule(lr){2-9}
 & \multirow{6}{*}{50\%} & 42 & 100 & 2.956785 & 27.236181 & 22.834646 & 25.159129 & 26.197655 \\
 &  & 42 & 150 & 2.653525 & 27.236181 & 22.834646 & 25.192630 & 26.214405 \\
 &  & 100 & 100 & 2.805155 & 26.666667 & 24.803150 & 24.656616 & 25.661642 \\
 &  & 100 & 150 & 2.805155 & 27.571189 & 24.409449 & 24.556114 & 26.063651 \\
 &  & 2026 & 100 & 3.184230 & 27.035176 & 25.590551 & 25.259631 & 26.147403 \\
 &  & 2026 & 150 & 3.032600 & 26.968174 & 25.590551 & 24.958124 & 25.963149 \\
\bottomrule
\end{tabular*}
\end{table}

\paragraph{OOD Recovery Consistency.}
The complete grid in Table~\ref{tab:app_global_recovery_complete} exposes the
seed-level values behind the recovery aggregates. Averaging the two transfer
paths within seed, Global RKU exceeds the unnormalized control at every
sparsity and checkpoint; the corresponding three-seed gains are
+2.21/+2.41 points at 30\%, +6.92/+6.97 at 40\%, and +5.62/+5.59 at 50\%
for 100/150 steps. The 100-to-150-step changes are small relative to these
method gaps.

\section{Additional Qwen-2.5-7B Comparisons}
\label{app:qwen_external_results}

\begin{table}[H]
\centering
\appendixtablestyle
\setlength{\tabcolsep}{2.8pt}
\caption{Qwen task-input calibration comparison. Entries are three-seed averages of accuracy (\%) across GSM8K, AQuA, and MathQA. The same calibration content and 16K-token budget are used by both methods within each row; $\Delta$ is Global RKU minus GISP. Q16K contains questions-only, whereas Full16K contains full question--solution text. Avg. rows are arithmetic means of Mean3 over the three sparsity levels.}
\label{tab:app_qwen_taskcal_match}
\begin{tabular*}{\linewidth}{@{\extracolsep{\fill}}clccc@{}}
\toprule
\textbf{Sparsity} & \textbf{Calibration} & \textbf{Global RKU} & \textbf{GISP} & $\boldsymbol{\Delta}$ \\
\midrule
\multirow{2}{*}{30\%}
& GSM8K-Q16K & $47.66\pm0.57$ & $\mathbf{48.64\pm0.54}$ & $-0.98$ \\
& GSM8K-Full16K & $50.12\pm1.01$ & $\mathbf{50.50\pm0.30}$ & $-0.37$ \\
\midrule
\multirow{2}{*}{40\%}
& GSM8K-Q16K & $\mathbf{45.96\pm1.55}$ & $42.17\pm1.27$ & $\mathbf{+3.79}$ \\
& GSM8K-Full16K & $45.49\pm1.61$ & $\mathbf{47.55\pm0.80}$ & $-2.06$ \\
\midrule
\multirow{2}{*}{50\%}
& GSM8K-Q16K & $\mathbf{39.35\pm1.21}$ & $30.74\pm0.92$ & $\mathbf{+8.61}$ \\
& GSM8K-Full16K & $40.12\pm0.99$ & $\mathbf{42.75\pm0.51}$ & $-2.63$ \\
\midrule
& Q16K Avg. & $\mathbf{44.32}$ & $40.52$ & $\mathbf{+3.81}$ \\
& Full16K Avg. & $45.24$ & $\mathbf{46.93}$ & $-1.69$ \\
\bottomrule
\end{tabular*}
\end{table}

\begin{table}[H]
\centering
\scriptsize
\setlength{\tabcolsep}{4pt}
\caption{Seed-42 calibration-corpus sensitivity of Global RKU at 40\%
structured sparsity. Every calibration view contains 32 sequences of length
512 (16,384 tokens), and all rows use the same downstream evaluator.
This single-seed diagnostic is reported separately from the later three-seed
Q16K/Full16K sweep.}
\label{tab:app_rku_corpus_sensitivity}
\begin{tabular}{llccc}
\toprule
\textbf{Model} & \textbf{Calibration} &
\textbf{GSM8K} & \textbf{AQuA} & \textbf{MathQA} \\
\midrule
\multirow{3}{*}{Qwen-2.5-7B}
& Wiki16K & 24.41 & 29.92 & 36.82 \\
& C4-16K & 3.34 & 24.41 & 30.22 \\
& GSM8K-Q16K & 60.50 & 27.95 & 44.46 \\
\midrule
\multirow{3}{*}{Llama-3.1-8B}
& Wiki16K & 1.52 & 22.44 & 27.50 \\
& C4-16K & 0.99 & 21.26 & 26.26 \\
& GSM8K-Q16K & 2.27 & 22.05 & 29.61 \\
\midrule
\multirow{3}{*}{Gemma-2-9B}
& Wiki16K & 13.42 & 27.17 & 35.64 \\
& C4-16K & 10.54 & 26.38 & 34.27 \\
& GSM8K-Q16K & 20.77 & 31.10 & 37.12 \\
\bottomrule
\end{tabular}
\end{table}

\paragraph{Task-Input Calibration.}
Table~\ref{tab:app_qwen_taskcal_match} gives the complete comparison underlying the 40\% result reported in the main text. With Q16K, the RKU--GISP difference is $-0.98$, $+3.79$, and $+8.61$ points at 30\%, 40\%, and 50\% sparsity. With Full16K, GISP is higher by 0.37--2.63 points across the same sparsity range. Averaged over the three sparsity levels, Q16K gives 44.32 Mean3 for RKU versus 40.52 for GISP (+3.81), whereas Full16K gives 45.24 versus 46.93 (-1.69). The two calibration views therefore produce different method orderings despite using the same token budget and structural seeds.

\paragraph{Recovery Dynamics.}
Figure~\ref{fig:qwen_recovery_grid} in the main paper places the GSM8K, AQuA,
and paired MathQA trajectories alongside the seed-level Step-0 normalization
audit. GSM8K exhibits the largest recovery response, AQuA is flatter, and paired
MathQA OOD performance changes little from 100 to 150 steps.

\paragraph{NIRVANA Protocol.}
NIRVANA is an external structured-pruning reference whose native search space includes MLP and GQA groups, as does native GISP, while RKU, Wanda-Struct, and Taylor-FO operate on coupled FFN channels. Its calibration protocol also differs from RKU. The same nominal sparsity therefore removes capacity from different structures, and its mathematical Step-0 values are reported under that native protocol rather than as a same-search-space ablation.

\paragraph{Wanda Seed Invariance.}
Under locked Wiki16K calibration, Wanda-Struct uses a deterministic score and fixed per-layer quota. Structural seeds 42, 100, and 2026 therefore retain the same mask, which explains the repeated 0.00 SD values in the main and appendix tables. Occasional 0.01--0.02 differences come from evaluator or numerical variation rather than a different pruning topology.

\section{Cross-Family Evaluation: Llama-3.1-8B and Gemma-2-9B}
\label{app:llama_complete_results}

\paragraph{Cross-Family Hardware.}
Unless stated otherwise, the Llama-3.1-8B and Gemma-2-9B cross-family pruning,
evaluation, and recovery experiments were run on NVIDIA L20 GPUs in BF16.
These runs are used for accuracy and recovery evaluation only; the physical
latency and memory measurements in Appendix~\ref{app:resource_accounting}
use the RTX 4090 protocol described above.

\begin{table}[H]
\centering
\appendixtablestyle
\caption{Llama-3.1-8B protocol summary for the three global methods used in the
cross-model comparison. Equal token counts do not imply equal calibration corpora or equal
pruning spaces.}
\label{tab:app_llama_protocol}
\begin{tabularx}{\linewidth}{@{}lYYY@{}}
\toprule
\textbf{Method} & \textbf{Calibration} & \textbf{Pruning space} & \textbf{Selection} \\
\midrule
Global RKU (Wiki16K) & WikiText, $32\!\times\!512=16{,}384$ tokens & all FFN channels & one-shot global top-$K$ \\
Global RKU (task-input) & GSM8K questions-only (primary) or full solutions (diagnostic), 16,384 tokens & all FFN channels & one-shot global top-$K$ \\
GISP (C4-16K) & C4, $32\!\times\!512=16{,}384$ tokens & MLP + GQA groups & loss-aligned iterative selection \\
GISP (task-input) & GSM8K questions-only or full solutions (Full16K diagnostic), 16,384 tokens & MLP + GQA groups & loss-aligned iterative selection \\
NIRVANA & BookCorpus, native candidate/anchor protocol & MLP + GQA groups & adaptive candidate selection \\
\bottomrule
\end{tabularx}
\end{table}

\paragraph{Cross-Family Protocol Scope.}
Table~\ref{tab:app_llama_protocol} separates RKU's FFN-only pruning space from the native MLP+GQA spaces used by GISP and NIRVANA. Equal calibration-token counts also do not make WikiText, C4, and GSM8K content equivalent. The Llama and Gemma results below test whether the RKU trends transfer across architectures under these method-specific protocols.

\begin{table}[H]
\centering
\appendixtablestyle
\caption{Llama-3.1-8B Step-0 comparison among global structured-pruning methods.
Values are mean $\pm$ sample SD over seeds 42, 100, and 2026. Task-input RKU
results are reported separately in Table~\ref{tab:app_llama_rku_taskinput_extension}.}
\label{tab:app_llama_step0_all}
\begin{tabular*}{\linewidth}{@{\extracolsep{\fill}}clccc@{}}
\toprule
\textbf{Sparsity} & \textbf{Method / calibration} & \textbf{GSM8K} & \textbf{AQuA} & \textbf{MathQA} \\
\midrule
\multirow{3}{*}{30\%}
& Global RKU (Wiki16K) & $\mathbf{2.50}_{\pm0.26}$ & $23.75_{\pm0.45}$ & $\mathbf{28.96}_{\pm0.25}$ \\
& GISP (C4-16K) & $0.00_{\pm0.00}$ & $22.44_{\pm1.72}$ & $27.43_{\pm0.63}$ \\
& NIRVANA (BookCorpus) & $1.84_{\pm0.04}$ & $\mathbf{25.20}_{\pm0.68}$ & $27.08_{\pm1.44}$ \\
\midrule
\multirow{3}{*}{40\%}
& Global RKU (Wiki16K) & $\mathbf{0.94}_{\pm0.18}$ & $23.23_{\pm0.39}$ & $\mathbf{26.06}_{\pm0.07}$ \\
& GISP (C4-16K) & $0.00_{\pm0.00}$ & $23.10_{\pm1.14}$ & $25.93_{\pm0.82}$ \\
& NIRVANA (BookCorpus) & $0.53_{\pm0.27}$ & $\mathbf{26.25}_{\pm0.60}$ & $22.91_{\pm0.32}$ \\
\midrule
\multirow{3}{*}{50\%}
& Global RKU (Wiki16K) & $\mathbf{1.62}_{\pm0.09}$ & $20.47_{\pm0.39}$ & $\mathbf{25.42}_{\pm0.15}$ \\
& GISP (C4-16K) & $0.00_{\pm0.00}$ & $22.57_{\pm1.38}$ & $24.76_{\pm0.32}$ \\
& NIRVANA (BookCorpus) & $0.35_{\pm0.29}$ & $\mathbf{23.36}_{\pm2.41}$ & $20.28_{\pm0.19}$ \\
\bottomrule
\end{tabular*}
\end{table}

\paragraph{Llama Comparison.}
Table~\ref{tab:app_llama_step0_all} compares the three global methods under
their reported calibration protocols. Wiki16K RKU exceeds GISP in eight of nine
direct task cells and exceeds NIRVANA on GSM8K and MathQA at all three
sparsities.

\begin{table}[H]
\centering
\appendixtablestyle
\caption{Complete Llama-3.1-8B RKU questions-only and full question--solution Step-0 results.
Values are mean $\pm$ sample SD over structural seeds 42, 100, and 2026. Mean3
is computed within seed over GSM8K, AQuA, and MathQA before three-seed aggregation.}
\label{tab:app_llama_rku_taskinput_extension}
\begin{tabular*}{\linewidth}{@{\extracolsep{\fill}}clcccc@{}}
\toprule
\textbf{Sparsity} & \textbf{Calibration} & \textbf{GSM8K} & \textbf{AQuA} & \textbf{MathQA} & \textbf{Mean3} \\
\midrule
\multirow{2}{*}{30\%}
& Q16K & $7.40\pm0.69$ & $21.92\pm1.26$ & $32.67\pm0.55$ & $20.67\pm0.24$ \\
& Full16K & $28.10\pm0.12$ & $25.33\pm1.27$ & $36.74\pm0.24$ & $30.06\pm0.47$ \\
\midrule
\multirow{2}{*}{40\%}
& Q16K & $1.92\pm0.61$ & $22.18\pm1.38$ & $30.28\pm0.60$ & $18.13\pm0.70$ \\
& Full16K & $19.54\pm1.53$ & $25.07\pm1.49$ & $35.48\pm0.51$ & $26.69\pm1.07$ \\
\midrule
\multirow{2}{*}{50\%}
& Q16K & $0.78\pm0.16$ & $21.78\pm1.20$ & $26.41\pm0.76$ & $16.33\pm0.23$ \\
& Full16K & $10.74\pm0.74$ & $24.15\pm1.20$ & $32.22\pm0.15$ & $22.37\pm0.21$ \\
\bottomrule
\end{tabular*}
\end{table}

\paragraph{Llama Questions-Only Calibration.}
Relative to the Wiki16K RKU rows in Table~\ref{tab:app_llama_step0_all}, Q16K
raises Llama Mean3 by +2.27, +1.39, and +0.49 points at 30\%, 40\%, and 50\%
sparsity. Task-level effects vary, while MathQA improves at all three sparsities.
Full16K is reported only as a diagnostic using full solutions outside the proposed
label-free operating point.

\begin{table}[H]
\centering
\appendixtablestyle
\caption{Llama-3.1-8B matched GISP task-input Step-0 results. All entries are
mean $\pm$ sample SD over seeds 42, 100, and 2026. Q16K uses questions-only;
Full16K uses full question--solution text.}
\label{tab:app_llama_gisp_taskinput}
\begin{tabular*}{\linewidth}{@{\extracolsep{\fill}}clcccc@{}}
\toprule
\textbf{Sparsity} & \textbf{Calibration} & \textbf{GSM8K} & \textbf{AQuA} & \textbf{MathQA} & \textbf{Mean3} \\
\midrule
\multirow{2}{*}{30\%}
& Q16K & $14.76\pm0.86$ & $22.31\pm0.60$ & $36.67\pm0.30$ & $24.58\pm0.56$ \\
& Full16K & $44.71\pm0.80$ & $28.74\pm0.79$ & $41.22\pm0.08$ & $38.22\pm0.25$ \\
\midrule
\multirow{2}{*}{40\%}
& Q16K & $5.08\pm0.67$ & $20.87\pm1.97$ & $34.20\pm0.57$ & $20.05\pm0.69$ \\
& Full16K & $37.65\pm1.12$ & $26.77\pm1.04$ & $38.40\pm0.49$ & $34.28\pm0.56$ \\
\midrule
\multirow{2}{*}{50\%}
& Q16K & $1.19\pm0.35$ & $22.31\pm2.31$ & $30.40\pm0.53$ & $17.96\pm0.83$ \\
& Full16K & $23.33\pm1.53$ & $25.07\pm1.14$ & $35.42\pm0.51$ & $27.94\pm0.70$ \\
\bottomrule
\end{tabular*}
\end{table}

\paragraph{Full-Solution Calibration.}
Table~\ref{tab:app_llama_gisp_taskinput} reports GISP under questions-only and
full question--solution calibration. The Full16K rows use solution traces and
therefore represent an operating point with richer supervision than label-free RKU;
we report them as a protocol diagnostic rather than as a label-free
comparison.

\begin{table}[H]
\centering
\appendixtablestyle
\caption{Llama-3.1-8B paired MathQA OOD recovery for the global-method comparison. Values are mean $\pm$ sample SD over structural seeds 42, 100, and 2026. Global RKU uses Wiki16K, GISP uses C4-16K, and NIRVANA uses its native BookCorpus protocol.}
\label{tab:app_llama_global_recovery}
\begin{tabular*}{\linewidth}{@{\extracolsep{\fill}}clcc@{}}
\toprule
\textbf{Sparsity} & \textbf{Method} & \textbf{OOD@100} & \textbf{OOD@150} \\
\midrule
\multirow{3}{*}{30\%}
& Global RKU & $\mathbf{31.13\pm0.28}$ & $\mathbf{30.94\pm0.09}$ \\
& GISP & $29.20\pm0.81$ & $29.10\pm0.54$ \\
& NIRVANA & $28.30\pm0.72$ & $28.20\pm0.82$ \\
\midrule
\multirow{3}{*}{40\%}
& Global RKU & $\mathbf{29.05\pm0.26}$ & $\mathbf{29.11\pm0.27}$ \\
& GISP & $27.78\pm0.49$ & $27.47\pm0.62$ \\
& NIRVANA & $25.15\pm0.13$ & $25.37\pm0.41$ \\
\midrule
\multirow{3}{*}{50\%}
& Global RKU & $\mathbf{27.82\pm0.35}$ & $\mathbf{27.72\pm0.09}$ \\
& GISP & $26.62\pm0.32$ & $26.54\pm0.30$ \\
& NIRVANA & $23.32\pm0.28$ & $23.53\pm0.22$ \\
\bottomrule
\end{tabular*}
\end{table}

\paragraph{Llama Recovery.}
Global RKU is above GISP and NIRVANA in all six paired OOD recovery cells in Table~\ref{tab:app_llama_global_recovery}. Relative to the direct MathQA Step-0 values in Table~\ref{tab:app_llama_step0_all}, the 150-step change for RKU is +1.98, +3.05, and +2.30 points at 30\%, 40\%, and 50\% sparsity. The corresponding GISP changes are +1.67, +1.54, and +1.78, while NIRVANA changes by +1.12, +2.46, and +3.25. Thus RKU shows the largest recovery increment at 30\% and 40\%, while NIRVANA has the largest increment at 50\%; RKU nevertheless remains highest in absolute OOD@150 accuracy at all three sparsities. Because the calibration protocols differ, this table is interpreted as a cross-family recovery check rather than a matched leaderboard.

\subsection{Gemma Direct and Task-Input Results}

\begin{table}[H]
\centering
\appendixtablestyle
\caption{Gemma-2-9B direct post-pruning global-method results with completed
Global RKU and GISP task-input sweeps plus the NIRVANA BookCorpus reference.
Values are mean $\pm$ sample SD over structural seeds 42, 100, and 2026. Q16K
uses questions-only and Full16K uses full question--solution text. NIRVANA uses
its native cross-protocol reference setting; no cross-protocol leaderboard
ranking is implied.}
\label{tab:app_gemma_step0_complete}
\begin{tabular*}{\textwidth}{@{\extracolsep{\fill}}clccc@{}}
\toprule
\textbf{Sparsity} & \textbf{Method / calibration} & \textbf{GSM8K} & \textbf{AQuA} & \textbf{MathQA} \\
\midrule
\multirow{6}{*}{30\%}
& Global RKU (Wiki16K) & $24.39\pm1.73$ & $30.05\pm0.23$ & $39.69\pm0.24$ \\
& Global RKU (GSM8K-Q16K) & $36.72\pm0.70$ & $30.05\pm0.23$ & $41.02\pm0.31$ \\
& Global RKU (GSM8K-Full16K) & $39.65\pm1.64$ & $29.13\pm0.79$ & $41.08\pm0.34$ \\
& GISP (GSM8K-Q16K) & $42.28\pm1.65$ & $34.51\pm0.23$ & $41.51\pm0.53$ \\
& GISP (GSM8K-Full16K) & $57.92\pm0.67$ & $37.01\pm1.04$ & $44.22\pm0.47$ \\
& NIRVANA (BookCorpus) & $20.65\pm2.10$ & $27.30\pm2.24$ & $35.30\pm0.48$ \\
\midrule
\multirow{6}{*}{40\%}
& Global RKU (Wiki16K) & $11.27\pm1.34$ & $29.40\pm0.60$ & $36.04\pm0.54$ \\
& Global RKU (GSM8K-Q16K) & $19.66\pm0.97$ & $30.45\pm0.60$ & $37.00\pm0.16$ \\
& Global RKU (GSM8K-Full16K) & $24.39\pm1.64$ & $27.56\pm0.68$ & $37.90\pm0.69$ \\
& GISP (GSM8K-Q16K) & $28.91\pm1.61$ & $30.18\pm0.45$ & $37.96\pm1.00$ \\
& GISP (GSM8K-Full16K) & $47.38\pm1.56$ & $32.02\pm1.64$ & $40.92\pm0.48$ \\
& NIRVANA (BookCorpus) & $4.57\pm0.37$ & $24.02\pm1.57$ & $29.27\pm0.14$ \\
\midrule
\multirow{6}{*}{50\%}
& Global RKU (Wiki16K) & $2.83\pm0.12$ & $24.80\pm0.39$ & $31.45\pm0.46$ \\
& Global RKU (GSM8K-Q16K) & $7.43\pm0.00$ & $27.56\pm1.36$ & $34.10\pm0.20$ \\
& Global RKU (GSM8K-Full16K) & $11.27\pm0.72$ & $27.69\pm0.91$ & $35.76\pm0.52$ \\
& GISP (GSM8K-Q16K) & $16.20\pm1.09$ & $27.17\pm0.68$ & $35.25\pm0.87$ \\
& GISP (GSM8K-Full16K) & $34.34\pm1.14$ & $31.36\pm0.60$ & $38.88\pm0.57$ \\
& NIRVANA (BookCorpus) & $0.61\pm0.30$ & $20.87\pm0.39$ & $23.71\pm0.19$ \\
\bottomrule
\end{tabular*}
\end{table}

\paragraph{Gemma Calibration Transfer.}
Table~\ref{tab:app_gemma_step0_complete} shows that Q16K improves RKU over
Wiki16K at every sparsity, with Mean3 gains of +4.55, +3.47, and +3.34 points
at 30\%, 40\%, and 50\%. The improvement appears across multiple tasks, with the largest gains on GSM8K: GSM8K rises by 12.33/8.39/4.60 points, MathQA by
1.33/0.96/2.65, and AQuA is unchanged at 30\% before improving by 1.05 and
2.76 points. GISP and Full16K show different task orderings, so this matrix
supports cross-family calibration sensitivity rather than a universal
leaderboard claim.

\clearpage
\section{Protocol-Conditioned Localization Diagnostics}
\label{app:protocol_conditioned_diagnostics}

\paragraph{Diagnostic Scope}
The following experiments separate retained structure from calibration-distribution effects. Replicated three-seed results are used for quantitative claims; the retention-credit interventions are reported as mechanism diagnostics with their protocol scope stated explicitly.

\subsection{Retention-Credit Diagnostic}
\label{app:credit_diagnostic}

Table~\ref{tab:app_credit_motivation} summarizes the controlled comparisons
underlying the retention-credit diagnostic in \secref{sec:experiments} and states
their protocol scope.

\begin{table}[t]
\centering
\appendixtablestyle
\caption{Controlled interventions separating layer-wise allocation from
channel-level retention. FANG and GISP retain identical or highly correlated
layer allocations while their exact FFN masks change substantially. For
GRASPrune, the reported mask-shift ratio compares calibration-domain variation
with within-domain train-order variation.}
\label{tab:app_credit_motivation}
\begin{tabular*}{\linewidth}{@{\extracolsep{\fill}}llll@{}}
\toprule
\textbf{Method}
& \textbf{Intervention}
& \textbf{Layer-wise similarity}
& \textbf{FFN-mask change} \\
\midrule
FANG \cite{yu2026fang}
& calibration corpus
& Spearman $1.000$
& Jaccard $0.674$--$0.696$ \\

GISP \cite{wang2026gisp}
& objective, same data
& Spearman $0.967$--$0.981$
& Jaccard $0.651$--$0.690$ \\

GRASPrune \cite{wang2026grasprune}
& calibration domain
& Spearman $0.938$--$0.946$
& $1.135\times$ train-order shift \\
\bottomrule
\end{tabular*}
\end{table}

\section{Implementation Correspondence}
\label{app:implementation_concordance}

Table~\ref{tab:app_rku_concordance} records the executable counterpart of each
Global RKU component in \secref{sec:methodology}. In the reported Global
RKU and unnormalized-control runs, the minimum retained ratio is fixed
to $\rho_{\min}=0.10$. After the initial model-wide top-$K$ selection, any block
below this floor is restored to the minimum width and the same number of
lowest-scoring eligible channels is removed from other blocks, preserving the
exact global channel budget.

\begin{table}[H]
\centering
\appendixtablestyle
\caption{Correspondence between the Global RKU definition and implementation.}
\label{tab:app_rku_concordance}
\begin{tabularx}{\linewidth}{@{}p{0.16\linewidth}p{0.30\linewidth}Y@{}}
\toprule
\textbf{Component} & \textbf{Methodology} & \textbf{Implementation} \\
\midrule
Probe target & $\mathcal E(x)=\|\mH_{\mathrm{post}}^{(L)}(x)\|_F^2$ & sum of squared entries of the model-returned post-RMSNorm final hidden state \\
Channel activation & down-projection input $\mY_{l,c}(x)$ & forward hook at the input of the down projection \\
Channel gradient & $\mG_{l,c}(x)=\nabla_{\mY_{l,c}(x)}\mathcal E(x)$ & full backward hook at the same input \\
Participation & $\Mean\!\left(|\mY_{l,c}(x)\odot\mG_{l,c}(x)|\right)$ & sum over equal batch/token counts \\
Down-projection norm & $w_{l,c}=\|\mW_{\mathrm{down}}^{(l)}[:,c]\|_2$ & column-wise Euclidean norm of the down projection \\
Global RKU
& $\mathcal S_{l,c}^{\mathrm{RKU}}
=w_{l,c}\mathcal A_{l,c}/Z_l$
& down-projection norm times participation normalized by
$\max\{\sum_j\mathcal A_{l,j},10^{-12}\}$ \\
Budget & $K=\round{(1-s)\sum_l d_{\mathrm{ff}}^{(l)}}$ & stable descending global selection with a 10\% minimum retained width per block \\
Structural unit & coupled gate/up rows and down column & one shared retained-index mask per FFN channel \\
\bottomrule
\end{tabularx}
\end{table}

The sum-versus-mean difference in the participation row is a constant
positive factor within a run because every channel is evaluated on the same
token count, so it does not alter Global RKU rankings. The reported implementation
uses the same score definition, global budget, and selection protocol as the
methodology above.

\end{document}